\documentclass{article}

\usepackage{arxiv}

\usepackage[utf8]{inputenc} % allow utf-8 input
\usepackage[T1]{fontenc}    % use 8-bit T1 fonts
\usepackage[hidelinks]{hyperref}       % hyperlinks
\usepackage{url}            % simple URL typesetting
\usepackage{booktabs}       % professional-quality tables
\usepackage{amsfonts}       % blackboard math symbols
\usepackage{nicefrac}       % compact symbols for 1/2, etc.
\usepackage{microtype}      % microtypography
\usepackage{lipsum}		% Can be removed after putting your text content
\usepackage{graphicx}
\usepackage{natbib}
\usepackage{doi}

\usepackage{lineno}
\usepackage{soul}
\usepackage{subfiles}
\usepackage{amssymb}
\usepackage{pdfrender}
\usepackage{float}
\usepackage{pbox}
\usepackage{amsmath}
\usepackage{array}
\usepackage{relsize}
\usepackage{bm}
\usepackage{multirow}

\title{Graph Neural Networks for Pressure Estimation in Water Distribution Systems}

\date{} 					% Or removing it

\author{
    {Huy Truong}\thanks{Both authors contributed equally to this work} \\
        Bernoulli Institute \\ 
        University of Groningen \\ 
        Groningen, The Netherlands \\
	\texttt{h.c.truong@rug.nl} \\
        \And
	{Andr\'{e}s Tello$^*$} \\
        Bernoulli Institute \\ 
        University of Groningen \\ 
        Groningen, The Netherlands \\
	\texttt{andres.tello@rug.nl} \\
\And
	{Alexander Lazovik} \\
        Bernoulli Institute \\ 
        University of Groningen \\ 
        Groningen, The Netherlands \\
	\texttt{a.lazovik@rug.nl} \\
	\And
	{Victoria Degeler} \\
        Informatics Institute \\ 
        University of Amsterdam \\ 
        Amsterdam, The Netherlands \\
	\texttt{v.o.degeler@uva.nl} \\
}

\renewcommand{\headeright}{GNNs for pressure estimation in WDNs.}
\renewcommand{\undertitle}{Submitted for publication in Water Resources Research}
\renewcommand{\shorttitle}{}

\usepackage{verbatim}

\begin{document}
\maketitle

\begin{abstract}
	Pressure and flow estimation in Water Distribution Networks (WDN) allows water management companies to optimize their control operations. For many years, mathematical simulation tools have been the most common approach to reconstructing an estimate of the WDN hydraulics. However, pure physics-based simulations involve several challenges, e.g. partially observable data, high uncertainty, and extensive manual configuration. Thus, data-driven approaches have gained traction to overcome such limitations. In this work, we combine physics-based modeling and Graph Neural Networks (GNN), a data-driven approach, to address the pressure estimation problem. First, we propose a new data generation method using a mathematical simulation but not considering temporal patterns and including some control parameters that remain untouched in previous works; this contributes to a more diverse training data. Second, our training strategy relies on random sensor placement making our GNN-based estimation model robust to unexpected sensor location changes. Third, a realistic evaluation protocol considers real temporal patterns and additionally injects the uncertainties intrinsic to real-world scenarios. Finally, a multi-graph pre-training strategy allows the model to be reused for pressure estimation in unseen target WDNs. Our GNN-based model estimates the pressure of a large-scale WDN in The Netherlands with a MAE of 1.94mH$_2$O and a MAPE of 7\%, surpassing the performance of previous studies. Likewise, it outperformed previous approaches on other WDN benchmarks, showing a reduction of absolute error up to approximately 52\% in the best cases.
\end{abstract}

% keywords can be removed

\section{Introduction}
% ANDRES

	\label{sec:introduction}
		State Estimation in Water Distribution Networks (WDNs) is a general problem that encompasses pressure and flow estimation, often using scarce and sparsely located sensor devices. WDNs management companies rely on such estimations for optimizing their operations. Knowing the state of the network at any given time enables water managers to perform real-time monitoring and control operations. The research community and practitioners working in this field have resorted for many years to the power of mathematical simulation tools to reconstruct an estimate of the system hydraulics \citep{fu2022role, garzon2022machine}. However, pure physics-based simulation approaches have to overcome the challenges of (i) data scarcity which translates to partially observable systems, (ii) high uncertainty introduced by the large number of parameters to configure, unexpected changes in consumers' behavior reflected in uncertain demand patterns, and noisy sensor measurements, and (iii) extensive manual configuration that requires expert knowledge and usually hinders model re-usability in a different WDN  \citep{wang2021probabilistic, fu2022role}. The challenges associated with physics-based modeling of WDNs have motivated researchers to investigate the usage of data-driven approaches, or a combination of both, to address the state estimation problem \citep{meirelles2017calibration, lima2018metamodel}. 
		
		Graph Neural Networks (GNNs) is a data-driven approach that has shown successful results in several estimation problems where data lies outside the Euclidean domain, and can be modeled as a graph. Since WDNs can be naturally modeled as a graph, GNNs can exploit the relational inductive biases imposed by the graph topology. As a result, GNNs have also attracted the attention of researchers in the field of WDNs. For example, \citep{tsiami2021cyber} used  temporal graph convolutional neural networks, a combination of Convolutional Neural Networks (CNNs) and GNNs, to extract temporal and spatial features simultaneously in a model to detect cyber-physical attacks in WDNs. \citep{zanfei2022novel} leveraged GNNs for burst detection algorithms. GNNs are also used for integrated water network partitioning and dynamic district metered areas \citep{fu2022graph}. In the context of Digital Twins of WDNs, a GNN-based model is used for Pump Speed-Based State Estimation \citep{bonilla2022digital}. Other works are more similar to ours and use GNNs for pressure estimation \citep{hajgato2021reconstructing, ashraf2023spatial}. 
  
        In this work, we focus on pressure estimation by leveraging both physics-based simulation models and GNN-based data-driven approaches. Our work proposes a number of research contributions. First, we propose an advanced data generation model to overcome the lack of data required for model training. Our method relies on a mathematical simulation tool, but it does not consider time-dependent patterns, producing a more diverse training dataset. In addition, we include some control parameters that remain untouched in previous works which contributes to data variety and avoids that uncertainties propagate due to model simplification errors \citep{du2018direct}. Second, our GNN-based estimation model is robust to unexpected sensor's location changes due to the proposed training strategy that relies on random sensor placement. Third, we propose a realistic evaluation protocol. Thus, our test set generation method considers real time-dependent patterns and additionally injects the uncertainties intrinsic to real-world scenarios. Finally, the proposed GNN-based model is equipped with generalization capabilities by design, and a multi-graph pre-training strategy allows to reuse the model for pressure reconstruction in different WDNs.

        As a consequence, our model reconstructed the junction pressures of Oosterbeek, a large-scale WDN in the Netherlands, with an average 1.94mH$_2$O absolute error, which represents an 8.57\% improvement with respect to other models. Similarly, our model outperformed previous approaches on other WDNs benchmark datasets. The highest improvement was seen for C-Town \citep{ostfeld2012battle} with an absolute error decrease of 52.36\%, for Richmond \citep{van2001methodology} an error deacrease of 5.31\%, and 40.35\% error decrease for L-Town \citep{Vrachimis2022}. In addition, our first attempt on model generalization shows that a multi-graph pre-training followed by fine-tuning helps to increase the model performance. In our case, the absolute error on Oosterbeek network reduced from 1.94 to 1.91mH$_2$O following our generalization strategy.
		
		% Our work stands out from those found in literature because our GNN-based estimation model addresses the three aforementioned challenges, found in physics-based approaches, all at once. First, we propose a different data generation model to overcome the lack of data required for model training. Second, we propose a realistic evaluation protocol that takes into account the uncertainties intrinsic to real-world scenarios. Finally, the proposed GNN-based model is equipped with generalization capabilities by design, and a multi-graph pre-training strategy allows to reuse the model for pressure reconstruction in different WDNs.
		
		The remainder of this document is as follows. Section \ref{sec:problem_statement} describes the issues that need to be addressed by pressure reconstruction models and defines the criteria to assess the model capabilities. Section \ref{sec:related_work} depicts the related work in the field, narrowed to GNNs for node-level regression tasks and how previous work on GNN-based pressure estimation satisfies the criteria defined in the Section \ref{sec:problem_statement}. The methodology is presented in Section \ref{sec:methodology}, including the data generation process, a detailed description of our model architecture, and the details of the proposed approach for model training and evaluation. Section \ref{sec:exp_settings} describes the setup of the experimental phase. It includes a description of WDNs benchmark datasets used in this work, the base model configurations, and the evaluation metrics. Section \ref{sec:experiments} describes all the empirical evaluations of our approach. First, the experiments on the main use case of this study, Oosterbeek WDN, are depicted. Then, the experiments towards model generalization are shown. Next, the performance of the proposed model on different benchmark WDNs is presented. This section concludes with an ablation study to identify the contribution of the different components of the model architecture. A discussion of the most salient findings are presented in Section \ref{sec:discussion}. Finally, the conclusions are presented in Section \ref{sec:conclusion}.

\label{sec:intro}

% a section describes problems/ usecases
\section{Pressure estimation in water distribution networks}

%reopen when this subsection is finished for the internal check
\subsection{Problem statement}
%Huy
%This section should begin with a subsection, where a formal task of pressure estimation is introduced. Also, generalisation is mentioned later, which is also never introduced. You can introduce it here as well.

%The need to fully understand essential hydraulic measurements and the role of state estimation. Limit the work to a particular measurement (i.e., pressure) and explain why
%issues that need to be addressed by pressure reconstruction models 
Hydraulic experts have managed water distribution networks using essential measurements such as flow, demand, and pressure. These measurements offer a comprehensive perspective of a water distribution network, forming a foundation for various supervisory tasks like forecasting, leak detection, and operational control. In this study, we narrowed down our work to estimate pressure due to the ease of meter installation and the more affordable price compared to flow ones \citep{zhou2019pressurebetter}. Nevertheless, to approximate a complete view of pressure in different locations in the water network, we use a data-driven model that can confront existing issues in practical water distribution networks.

%Network has limited observation nodes; thus, pressure estimation leverages these existing measurements to estimate the values of other nodes;
A real-life water network can include thousands of junctions indicating water outlets, customers, and pipe interactions. However, only some junctions are sensor-equipped and well-maintained due to infrastructural limits and privacy concerns. Thus, they raise the need for more data and observable sensors to train a pressure estimation model in a high-quality manner. Specifically, during training, the model -- typically structured with deep learning architectures as its backbone -- learns to estimate the pressure at all unknown junctions within the network, relying on measurements from only a limited number of sensors. This approach recalls a typical semi-supervised learning problem with more considerations in the deployment context.

%extended issue is a generalized ability of models.
The application context is about what and when the trained model should be applied. Generally, a model is often associated with a unique water network and fixed sensors previously seen during training. Also, the training environment may exclude noisy, uncertain conditions that could affect the model's decision-making. In other words, these challenges result in worse model performance when faced with unfamiliar network topologies or uncertain situations. Consequently, model retraining is inevitable, albeit such training is an expensive and unsustainable approach. This concern enhances the necessity of the generalization ability of pressure estimation models, which needs to be addressed in prior research. Before addressing this research gap, we will first delve into the specific problem within water networks and lay out the criteria necessary for a robust pressure estimation model.

%unknown distribution, limits of hydraulic simulation models, 
\subsection{Partially-Observable Data and Realistic Model Evaluation}
\label{sec:phenomenon}
%ANDRES 
Water Distribution Networks domain is characterized by partial-observability due to the limited sensor coverage. This imposes an additional challenge because the reconstruction models need to be trained on fully-observable network operation snapshots. The common approach to overcome this limitation is to rely on mathematical hydraulic simulation tools, e.g. EPANET \citep{rossman1999epanet}, WNTR \citep{klise2018overview}, to generate full-views of the network operation and use them for model training \citep{hajgato2021reconstructing, luxing2022gnn, ashraf2023spatial, zhou2023testing}. 

Although the hydraulic simulations solve the lack of training data for the reconstruction models, the remaining challenge is how to create a valid and reliable evaluation protocol and the data used for it. Sampling from the data generated from the simulation models and splitting them into training and test sets is not enough. Ideally, the assumption behind machine learning models is that the training data is ruled by the exact same distribution of the data on which the model will be evaluated. However, having absolute control over the data generation process and meeting such a perfect match between both distributions is unrealistic and the assumption is violated under real-world conditions \citep{bickel2007discriminative, hendrycksG17, fang2022out}. Thus, the prediction models should be robust to distribution shifts between training and testing samples, i.e., be able to generalize to out-of-distribution (OOD) data \citep{farquhar2022what}. 

\begin{figure}[b]
    \centering
    \noindent\includegraphics[width=0.95\textwidth]{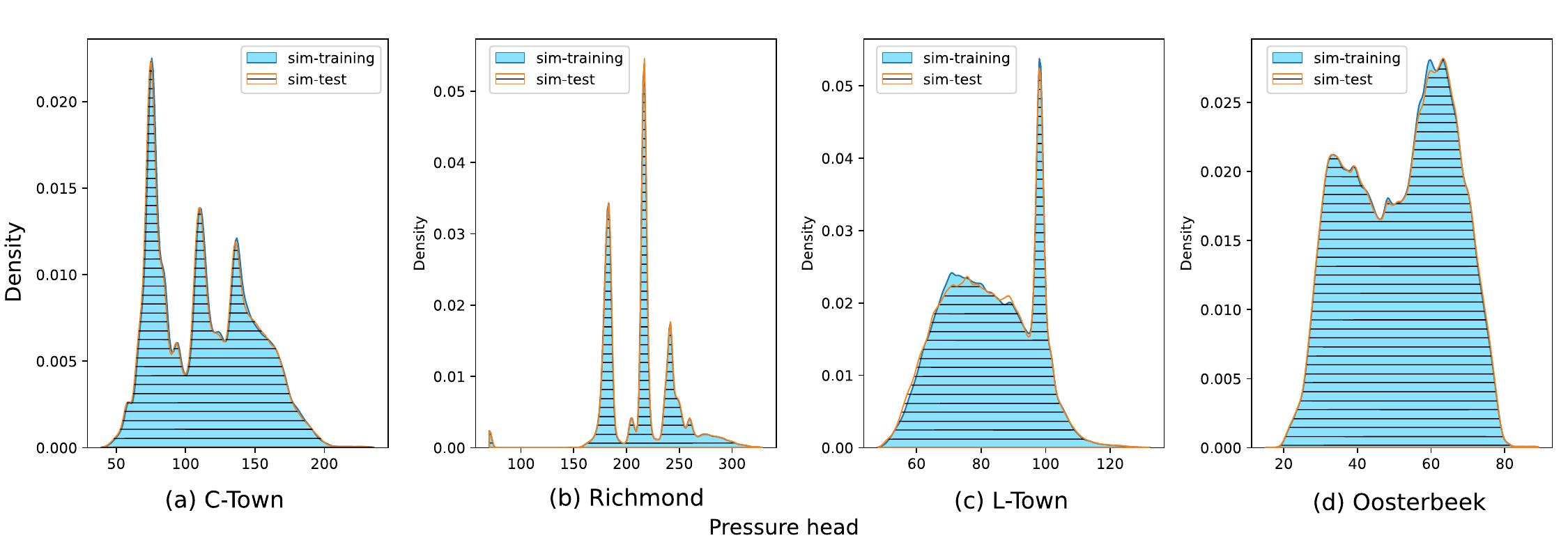}
    \caption{Density Distribution of training and test sets in C-Town, Richmond, L-Town and Oosterbeek WDNs, generated by the Hydraulic Simulation tool EPANET.}
    \label{fig:identical_data_dist}
\end{figure}

We observed that the data distribution of the training and test sets created by the simulation models are identical, which is unnatural in practice. The density distributions of training and test sets from different WDNs, generated by the Hydraulic Simulation tool EPANET, are shown in Figure \ref{fig:identical_data_dist}. In this case, the simulation's control parameters,  e.g. reservoir total heads, junction demands, pump speed, were randomly adjusted for every run. Nonetheless, as evident from the image, the distributions of the training and test sets created by the hydraulic simulations are identical in all examples. In this case, evaluating the reconstruction models on data generated by the same algorithm is simply evaluating the ability of the model to reconstruct the signals already seen during the training process. 

In our work we propose a realistic test set generation process that relies on time-based demand patterns. In addition, Gaussian noise is injected before the simulation to mimic the uncertainty intrinsic to real-world scenarios. Combining time-based demand patterns and noise injection allows to create realistic scenarios to evaluate the ability of the models to generalize to OOD data, with visible differences in density distribution between training and test sets.

\subsection{Criteria for model assessment}
%Huy
\label{sec:crit}

The out-of-distribution problem is a persistent challenge in mathematical simulations, originating from uncertainty and variability in hydraulic parameters, such as consumer demand, pipe roughness, and material aging attributes. Modeling and monitoring these values are complex and costly. This difficulty arises exponentially in sophisticated cases, such as fluids in curved pipes, lack of measurements, or when exterior reasons cause unforeseen effects on the network \citep{campos2021impact}. In addition, the generalization ability is weak because the hydraulic simulations cannot be applied to an unseen water network. Thus, they cannot deal with such problems and lag behind their standard capability.  

In light of the limitations of conventional simulations, previous studies have proposed using more efficient surrogate models (which we will discuss in Section \ref{sec:related_work}). However, it is critical to note that they overlooked the OOD problem. Thus, it essentially motivates a list of criteria that indicate the favorable capabilities of a surrogate model addressing the pressure estimation task on water distribution networks while taking into account the OOD, generalizability, and flexibility as follows. 

\begin{enumerate}
    \setlength\itemsep{1em}
    \item[] $\boldsymbol{(\mathbb{C}1)}$ The surrogate model should have topology awareness to effectively solve the pressure estimation task. In addition, it must be able to perform seamlessly on any Water Distribution Network, regardless of whether its topology is observable during training. This is an important aspect of generalizability, making it more useful in practice.

    \item[] $\boldsymbol{(\mathbb{C}2)}$ The surrogate model should be adaptable to various contextual circumstances. Adjusting a variety of sensor measurements can be a typical example. This criterion allows for model flexibility when new measurement meters are introduced. In addition, it counters situations where one or more sensors are deactivated for maintenance purposes.

    \item[] $\boldsymbol{(\mathbb{C}3)}$ To capture the OOD problem, model robustness should be taken into account, especially in the evaluation phase. The uncertainty inherited from the real world can yield noise in observations, including data transmission and discrepancy of hydraulic parameters between simulated and actual water networks. Many existing approaches do not address these issues, as they are often tested in well-simulated and noise-free cases that do not account for uncertainty.
\end{enumerate}

Satisfying all factors simultaneously is difficult. For this reason, the provided list is used to evaluate the state-of-the-art methods for estimating pressure in the next section. Then, we will introduce our solution that fulfills all three criteria on the list.
Our empirical experiments have shown that the suggested model outperforms other baselines, even when taking into account its parameter complexity.

%\quickwordcount{problem_statement} 
 %Huy, Andres
\label{sec:problem_statement}

\section{Related Work}
\label{sec:related_work}

\subsection{GNNs for node-level regression task}

\citep{wu2021survey} categorized GNN based on their purposes into graph-level, link-level, and node-level tasks. These categories indicate the versatility and primary focus of GNN to provide outcomes across various domains. For instance, graph-level and link-level have been employed in domains such as chemistry \citep{reiser2022chemistry}, bioinformatics \citep{nguyen2020bioinfo}, and recommendation systems \citep{chen2020kgc}. On the other hand, the node-level category has been dominated by node classification tasks. One illustrative example is in the field of physics, where GNN can predict the probability that an individual particle is associated with the pileup part of an event\citep{Shlomi2021physics}. In finance, loan fraud detection within consumer networks is a popular example of a node classification task \citep{xu2021finance}.

As an influence of prevalent node classification tasks, well-known GNN architectures have been developed to excel in this domain \citep{defferrard2016chebnet, chen2020gcn2, velickovic2018gat}. This focus led to a relative lack of attention in node regression tasks and caused ambiguity regarding their effectiveness in handling continuous values within the node-level regime. Hence, in this paper, we aim to bridge the research gap and explore the potential of these methods in addressing node regression challenges.

While some studies have started exploring node-level regression in specific domains, such as traffic \citep{derrow2021trafficforecasting} and recommendation systems \citep{ying2018node}, the water domain remains relatively unexplored. With this in mind, our primary objective is to compare popular approaches in a regression task known as pressure estimation and to introduce our cutting-edge GNN architecture designed for this purpose. It could open new avenues for GNN applications, especially in water management.

\subsection{State Estimation with GNNs}
State Estimation plays a critical role as a fundamental process that provides adequate information for WDN management, monitoring, and maintenance, such as leak localization \citep{mucke2023sametestway}, optimal control \citep{martinez2007optimalcontrol} and cyber-attack detection \citep{riccardo2018cyberattackonctwon}. Recent works have started to study GNNs when they outperformed classical models in graph-related tasks, especially in pressure estimation \citep{hajgato2021reconstructing}. Generally, GNN attempts to predict all pressure values at nodes using limited historical sensor values. For an overview, we list the most important works and evaluate them against the predefined criteria.

\citep{hajgato2021reconstructing} is the first work that proposes to train a GNN model named ChebNet on a well-defined synthetic dataset. $\boldsymbol{(\mathbb{C}2)}$ are satisfied because the authors trained the model on various snapshots concerning different sensor locations. In contrast, achieving $\boldsymbol{(\mathbb{C}3)}$ is vague as all reports were based on time-irrelevant and synthetic data. In addition, the model cannot deal with the generalization problem due to the limitation of spectral-based models \citep{zhang2019graph}. Concretely, each spectral model is trained on and linked to a particular topology, so using a single model on diverse WDNs is impractical. Hence, it fails to satisfy $\boldsymbol{(\mathbb{C}1)}$. 

\citep{ashraf2023spatial} improved the above work on historical data. In this case, working with a spatial-based GNN can solve the generalization problem, so it is possible to satisfy $\boldsymbol{(\mathbb{C}1)}$. In addition, testing on noisy time-relevant data could be seen as an uncertainty consideration
%, although this solution is inefficient due to the enormous model size(nearly 2.5 million parameters)
. However, this work heavily depends on historical data generated from a pure mathematical simulation engaging with ``unchanged" dynamic parameters (e.g., customer demand patterns). In practice, this approach does not apply to the cases where those parameters are prone to error or unknown \citep{kumar2008state}. Hence, we consider that it weakly satisfies $\boldsymbol{(\mathbb{C}3)}$. However, the author fixed sensor positions during training, which could negatively affect the model observability to other regions in the WDN. For this reason, stacking very deep layers that increase model complexity is inevitable to ensure the information propagation from far-away neighbors to fixed sensors \citep{Barcelo2020NeedsNLayersToProp}. Additionally, retraining the model is mandatory whenever a new measurement is introduced, which has detrimental effects on its flexibility and scalability. Thus, the model violates $\boldsymbol{(\mathbb{C}2)}$.

Note that we exclude the heuristic-based methods as they do not consider the topology in decision-making. Also, several graph-related approaches \citep{kumar2008state, luxing2022gnn} have existed in this field. However, they accessed historical data, and neither attempted to solve the task in a generalized manner. Alternatively, we assume that prior knowledge (i.e., historical data) is unavailable. In this work, we delve into the capability of GNNs in a general case, in which the trained model can be applied to any WDN and any scenario.
%\quickwordcount{related_work} 
 %Huy

\section{Methodology}
\label{sec:methodology}

\subsection{Water network as graph} % Huy

%Huy
A water distribution network is a complex infrastructure that provides safe and reliable access to clean water for individual usage. Thus, it is crucial to ensure it properly functions and sustainably meets the needs of the serving community. This monitoring process starts with gathering information from data streams captured by sensors installed in the network. For notation, we define a finite segment of the data stream as a scenario. Each scenario is divided into \textit{snaphots}, preserving the network state at a particular timestamp.

Mathematically, a \textit{snapshot} is represented as a finite, homogeneous, and undirected graph $\mathcal{G} = \{\mathcal{X}, \mathcal{E}, \mathcal{A}\}$ that has $N$ nodes and $M$ edges. Edges represent pipes, valves, and pumps, while nodes can be junctions, reservoirs, and tanks. The nodal features are stored in the matrix $\mathcal{X} \in \mathbb{R}^{N\times d_{node}}$, where $|\mathcal{X}| = N$ and $d_{node}$ is the number of feature channels. In this work, \textit{pressure} is the unique node feature because it is recognized as the most vital stable factor in monitoring the WDN \citep{christodoulou2018wdndefinition} and aligns with prior research\citep{hajgato2021reconstructing, ashraf2023spatial}. Thus, from now, we refer $\mathcal{X}$ to a \textit{pressure} matrix, and the feature dimension $d_{node}$ is fixed to 1.

%for consistency in previous work (i.e., $d_{node} = 1$), but not limited to the head, flow, or other measurements unless they are all homogeneous. The heteorophily discovery is also interesting, which we leave for future work.

$\mathcal{E} \in \mathbb{R}^{M \times d_{edge}}$ is an edge feature matrix, where $|\mathcal{E}| = M$ and $d_{edge}$ is the edge dimension. Depending on a particular model, we set $d_{edge}$ to $0$ if unused or $2$, which indicates pipe lengths and diameters are supported. %It is worth noting that this information can be gathered from static parameters corresponding to a Water Distribution Network stored in an INP file.
The node connection is represented in an adjacency matrix $\mathcal{A} \in \mathbb{R}^{N \times N}$, where $a_{ij}=1$ means node $i$ and $j$ are connected by a link, whose edge attribute $e_{ij} \in \mathcal{E}$, and $a_{ij}=0$ for otherwise.

Observing accurate pressure $\mathcal{X}$ for an entire water network is challenging due to partial observability. Hence, we rely on a physics-based simulation model to construct synthetic pressure as training samples for the surrogate model. Concretely, the simulation takes a range of simulation parameters, including static parameters, such as nodal elevation and pipe diameter, and dynamic parameters, like junction demands and tank settings. Then, it solves a hydraulic equation to estimate the \textit{pressure} and \textit{flow} at unknown nodes in a Water Distribution Network \citep{simpson2011equation}. Note that in our case, as we tackle a multi-topology problem, we flexibly compute head losses modeled in pipes using various formulas, such as Hanze-Williams, Daisy-Wechbach, and Chezy-Manning. For more details on solving hydraulic optimization, we refer the reader to the EPANET engine, which serves as our default mathematical simulation \citep{rossman1999epanet}. 

Despite the usability of conventional simulations, they demand a manual calibration process to stay synchronized with the actual physical water network. Also, they lack efficiency and suffer from the OOD problem mentioned in Section \ref{sec:problem_statement}. In light of these limitations, we adopt a strategic alternative. In particular, we merely leverage a simulation model to generate synthetic samples once.
Subsequently, these synthetic samples serve as the training data for our calibration-free surrogate model. The trained model can infer the pressure of any water network in the deployment. In the following section, we focus on the first stage, where we construct the training dataset using the conventional simulation.

%\quickwordcount{problem_statement} 

\subsection{Dataset creation}% Huy

%assumption should be already stated in the introduction or the 2nd section

Throughout this paper, GNNs take WDN \textit{snapshots} as input features. In particular, each \textit{snapshot} provides a global view of a WDN graph representing concrete pressure values at an arbitrary time. Additionally, it contains topological information (e.g., nodal degree, node connectivity, and edge attributes) from a corresponding water network. We denote as a \textit{clean snapshot} the one that describes an instantaneous pressure state without any hidden information. In contrast, a \textit{masked snapshot} portrays a partially observable network in which the target feature known as pressure is mostly undetermined except for a small number of metered areas.

The conventional generation requires temporal patterns to create a set of \textit{clean snapshots}. A pattern records a time series of a specific simulation parameter, such as customer demand or pump curve, in a fixed period. In other words, such a series is periodic and bound in common scenarios.
%In other words, this trend is repetitive and bounded into a pre-defined temporal boundary.
However, it is not guaranteed that these patterns include all events, and real-world data is highly volatile. For example, a model trained on data created from past patterns can fail to estimate the pressure of a WDN during the long-term COVID-19 pandemic due to the unexpected sudden change in water consumption that was never found in such patterns \citep{campos2021impact}. Furthermore, the number of available patterns is seldom provided or partially accessible due to privacy-related concerns, especially in public benchmark WDNs. For this reason, they are often repetitively overused in modeling large-scale water networks where the number of nodes is exponential compared to the required patterns. This significantly impacts dataset diversity and, therefore, limits the model capability to satisfy criteria $\boldsymbol{(\mathbb{C}3)}$.

Before proposing our alternative solution, we review existing generation methods to analyze the effect of time-series patterns on simulation parameters. Generally, two existing options include time-dependent and sampling-based ones. Table \ref{tab:param_selection} indicates the existing methods for selecting and altering dynamic parameters. The underlying simulation still plays a crucial role in creating \textit{clean snapshots} given an arbitrary set of parameters. Still, each method has a specific selection and adjustment of dynamic parameters with respect to a design space.

\begin{table}[b]
\caption{\textbf{The selection of dynamic parameters between conventional simulations and sampling-based generations.} Parameters marked with a check can exhibit varying values, whereas others remain constant throughout the generation process. Note that the dynamic parameter selection also depends on component availability in a particular network and dataset creation stability to prevent abnormal results. }
\centering
\label{tab:param_selection}
\resizebox{0.98\columnwidth}{!}{%p{0.15\linewidth}
    \begin{tabular}%${@{\quad}l c c c c c c c@{\quad}}
    {@{\quad}l *{8}{>{\centering\let\newline\\\arraybackslash\hspace{0pt}}p{1.4cm}} @{\quad}}
    \toprule
         \pbox{1.5cm}{Dataset Creation} & \pbox{1.4cm}{Reservoir Total Heads} & \pbox{1.3cm}{Junction\\Demand} & \pbox{1.4cm}{Pump\\Speed} & \pbox{1.4cm}{Pump\\Status}  & \pbox{1.4cm}{Tank\\Levels} & \pbox{1.4cm}{Valve\\Settings} & \pbox{1.4cm}{Valve\\Status}  & \pbox{1.4cm}{Pipe\\Roughness}   \\
    \midrule
        \citep{rossman1999epanet} & \checkmark & \checkmark & \checkmark$^{a}$
        &   &   &   &  & \\
        
        {\citep{hajgato2020datagen}}&  & \checkmark & \checkmark & \checkmark & \checkmark  &  & &\\
        
        Our& \checkmark & \checkmark & \checkmark & \checkmark & \checkmark & \checkmark & \checkmark & \checkmark
        \\\bottomrule 
    \multicolumn{7}{l}{$^{a}$ \textit{Pump speed} is implicitly adjusted by a \textit{pump curve} pattern. }
    
    \end{tabular}
}
\end{table}

The conventional simulation method, EPANET \citep{rossman1999epanet}, operates time-dependently and relies primarily on fixed patterns. Excessive use of these patterns results in temporal correlations among snapshots, primarily due to their inherent seasonal factors. This issue becomes inevitable, especially in large-scale networks, where numerous unmeasurable nodes require pattern assignments to complete a simulation process. Consequently, this leads to information leakage between snapshots within the same scenario (see after-splitting data distribution in training and testing sets in Figure \ref{fig:identical_data_dist}). 

Alternatively, \citep{hajgato2020datagen} eliminate time patterns and consider a single snapshot as an instantaneous scenario. This way is more delicate to provide more observations for data-hungry models. However, \citep{hajgato2020datagen} focus only on pump optimization, so half of the listed parameters remain untouched. 

Both available generations assume the remaining parameters are deterministic and unchanged. Nevertheless, these parameters (e.g., pipe roughness) can be critical factors affecting the simulation result \citep{zanfei2023shall}. Thus, neglecting any of these parameters can restrict the model in learning representations of WDN snapshots.

Intuitively, we consider a comprehensive modification of all dynamic parameters as a data augmentation to ensure the simulation quality and address the generalization problem. Our main objective is to design a sufficient search space to provide different pressure views from flexible sensor positions. This approach helps alleviate the data-hungry issue when training deep learning models and benefits model robustness thanks to the augmented data space \citep{cubuk2020randaugment}. 

\begin{figure}[h]
    \centering
    \noindent\includegraphics[width=0.8\textwidth]{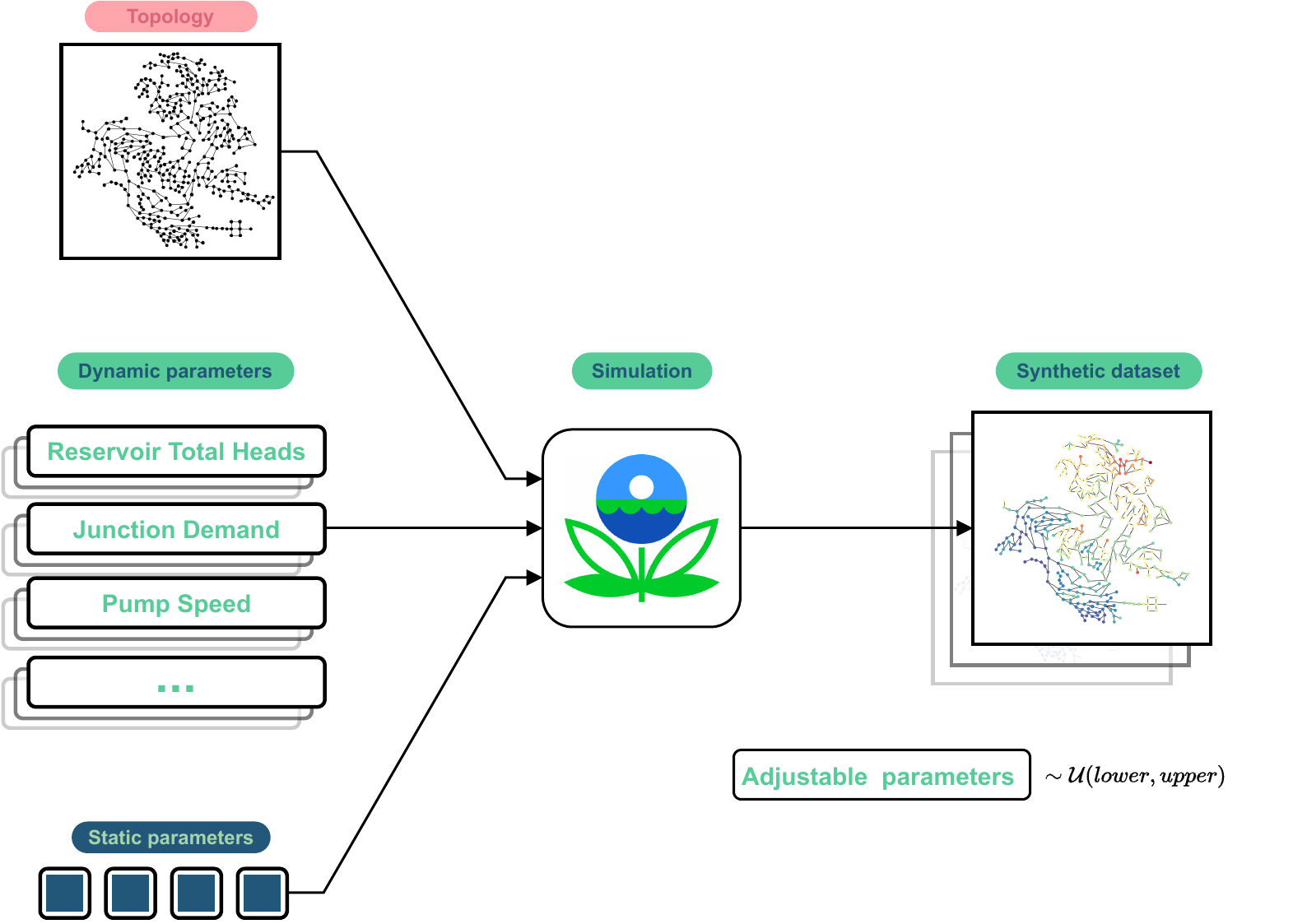}
    \caption{\textbf{Our data generation approach.} Dynamic parameters are sampled from a uniform distribution and passed to the mathematical simulation with static values and WDN topology. The result is a synthetic dataset containing legit snapshots whose pressure range should be close to reality.}
    \label{img:gen_visual}
\end{figure}

In particular, we adopt a brute-force approach to explore the full range of available dynamic parameters. To ensure the simulation quality, we exclude parameter sets producing pressure ranges exceeding practical limits. Subsequently, our generation takes these sets of dynamic parameters, an unchanged static set, and the topology of a particular water network to generate a single snapshot using the conventional simulation (refer to Figure \ref{img:gen_visual}). Note that it only performs a single simulation step that removes the essence of temporal patterns. This process outcome is a set of distinct immediate pressure states, which are more versatile and independent in time. In contrast to the classical usage, this approach eliminates the temporal correlation concern and leverages all dynamic parameters to generate training samples designed to cover the entire input space.

\subsection{Model Architecture}% Huy

The model is expected to learn a graph representation from existing known signals to estimate unknown pressures. After this, given a deterministic topology, we need an approach to spread local representations from meter nodes to distant neighbors efficiently. We first recap Message Passing Neural Networks (MPNN) \citep{gilmer2017mpnn}, the generic framework for spatial GNNs. Then, we discuss Graph Attention Network (GAT) \citep{velickovic2018gat} as one of our fundamental components. In light of this, we propose \textit{GATRes} as a principal block and devise the overall architecture illustrated in Figure \ref{img:architect}.

\subsubsection{Preliminaries}

Considering a GNN as a series of stacked layers, MPNN describes a specific layer to transform previous representations to successive ones using message propagation. 
We omit the layer index for simplicity and denote representations of a target node $i$ as $x_{i}$. Noticeably, the first representations are input features known as pressure values. Then, the output representations of a consecutive layer are computed as follows:

\begin{equation}
\label{eqa:msgpassing}
    z_i = \textrm{UPDATE}\left(x_i, \bigoplus_{j \in \mathcal{N}(i)}  \textrm{MSG}(x_j)\right)
\end{equation}

where $z_i$ is the corresponding output of the target node $i$,  $\mathcal{N}(i)$ denotes the 1-hop neighbors, $\textrm{MSG}$ and $\textrm{UPDATE}$ are differential functions describing messages received from neighbors and the way to update that information concerning its previous representations respectively.

$\bigoplus$ is a differentiable, permutation-invariant function, ensuring the gradient flow backward for model optimization and addressing concerns related to node ordering \citep{gilmer2017mpnn}. This function plays a critical role in aggregating neighbor messages into the target one. 

Depending on the task-specific purpose, numerous ways exist to define the message aggregator, such as mean, max, sum \citep{xu2019gin}, or Multilayer Perceptron \citep{zeng2020graphsaint}. Ideally, $\bigoplus$ is designed to propagate messages from surrounding nodes in a sparse fashion, which only matters to non-zero values. Thus, this scheme efficiently scales when dealing with enormous graphs and economically saves the memory allocation budget. 

Next, we explain GAT in view of a target node $i$. Concretely, GAT focuses on the intermediate representation relationship between the target node $i$ and one of its 1-hop neighbors $j$. If a node pairs with itself, it forms a self-attention relationship. Hence, we strategically establish a virtual self-loop link in every node to put weights between itself representations compared to the aggregated ones from the neighborhood. Mathematically, we can rewrite the GAT formula according to Equation \ref{eqa:msgpassing} as:
\begin{equation}
\label{eqa:gat}
    z_i = \overset{H}{\underset{h}{\bigg\vert\bigg\vert}} \underset{j \in \mathcal{N}(i) \cup \{i\}}{\mathlarger{\sum}} \alpha^h_{ij} \theta x_j = GAT(x_i)
\end{equation}
%concat: \overset{H}{\underset{h}{\Bigg\vert\Bigg\vert}}
%avg : \frac{1}{H}  \overset{H}{\underset{h=1}{\mathlarger{\sum}}}
where $H$ is the number of heads, $\vert\vert$ is a concatenation operator, $\theta \in \mathbb{R}^{d_{in} \times d_{out}}$ is the layer weight matrix with $d_{in}$ and $d_{out}$ that are the input and output representation dimensions, respectively. For each head $h$, the attention coefficient $\alpha$ is computed as: 

\begin{equation}
\label{eqa:gatalpha}
    \alpha_{ij} = \textrm{softmax}(\sigma(a^T [\theta x_i \vert \vert \theta x_j]))
\end{equation}
where $\textrm{softmax}(x) = \frac{e^{x_i}}{\sum_{j \in \mathcal{N}(i) \cup \{i\}} e^{x_j} }$ is used to compute the important score between the target node $i$ and a neighbor $j$. Before calculating $\textrm{softmax}$, the concatenation of both nodal representations is then parameterized by learnable weights $a \in \mathbb{R}^{1 \times 2d_{out}}$ and passed through a non-linear activation function $\sigma(.)$ (e.g., ReLU, GELU, or LeakyRELU \citep{xu2015empirical}).

Inspired from \citep{vaswani2017attention}, GAT leverages multiple heads to perform parallel computation and produce diverse linear views. In the original work, those head views should be joined to ``merge'' all-in-one representations, thanks to a linear layer mapping the concatenated heads. However, the conventional approach~\citep{velickovic2018gat} is to stack numerous concatenated GAT layers sequentially (hence, without any head joint) except for the last layer, where a final mean view is computed but only for the final logit in a classification task. The postponed head joining could preserve irrelevant views in the consecutive layer that double the detrimental effects of the nodal feature sparsity due to the high masking rate in an unsupervised setting. In other words, irrelevant head views quickly saturate the impact of final nodal presentations and accelerate the smoothing process (that is, oversmoothing \citep{chen2020oversmoothing}). Extra propagation layers are helpless because they worsen the situation. Thus, we hypothesize that merging head views could complete the original design and suppress unrelated information. Intuitively, whether to apply a linear transformation to the concatenation head view as in \citep{vaswani2017attention} or merely take an average of head representations arises.

% After a single GAT, the nodal dimension expands to H times due to the concatenation of the head and plays a role as the new node features for the next propagation layer. Intuitively, we can leverage the next layer to perform head merging and neighbor propagation simultaneously. In particular, we reuse the attention mechanism to weigh head views and fuse them into a uniform representation. 

\subsubsection{GATRes}
Alternatively, we propose using an additional GAT layer to evaluate head views generated from the previous one. We name this approach as \textbf{GAT} with \textbf{Res}idual Connections (GATRes).
Mathematically, we define our \textit{GATRes} as follows:

% \begin{equation}
%     z_i = x_i + \frac{1}{|\mathcal{N}(i)+1|} \underset{k \in \mathcal{N}(i) \cup \{i\}}{\sum} \beta_{ik} \Psi \Biggl( \overset{H}{\underset{h}{\big\vert\big\vert}} \underset{j \in \mathcal{N}(k) \cup \{k\}}{\mathlarger{\sum}} \alpha^h_{kj} \Theta x_j \Biggl)_k
% \end{equation}

% \begin{eqnarray}
%     &h_i = GAT(x_i; \alpha, \Theta)\\
%     &z_i = x_i + \frac{1}{|\mathcal{N}(i)+1|} \underset{k \in \mathcal{N}(i) \cup \{i\}}{\sum} \beta_{ik} \Psi h_k = x_i + \frac{1}{|\mathcal{N}(i)+1|} \underset{k \in \mathcal{N}(i) \cup \{i\}}{\sum} GAT(x_k; \beta, \Psi)
% \end{eqnarray}

% \begin{eqnarray}
%     &z_i = GAT(x_i; \alpha, \Theta)\\
%     &z_i = x_i + \frac{1}{|\mathcal{N}(i)+1|} \underset{j \in \mathcal{N}(i) \cup \{i\}}{\sum} GAT(z_j; \beta, \Psi)
% \end{eqnarray}

\begin{eqnarray}
    &z_i = x_i + \frac{1}{|\mathcal{N}(i)+1|} \underset{j \in \mathcal{N}(i) \cup \{i\}}{\sum} GAT(GAT(x_j; \alpha, \Theta); \beta, \Psi)
\end{eqnarray}
where, attention coefficients $\alpha \in \mathbb{R}^{N\times H}$ computed in Equation \ref{eqa:gatalpha} and learnable weight matrix $\Theta \in \mathbb{R}^{d_{in}\times Hd_{out}}$ belong to the first GAT. Identically, $\beta \in \mathbb{R^N}$ and $\Psi \in \mathbb{R}^{Hd_{out} \times d_{out}}$ are from the second GAT but different in shape. 

\begin{figure}[h]
    \centering
    \noindent\includegraphics[width=0.9\textwidth]{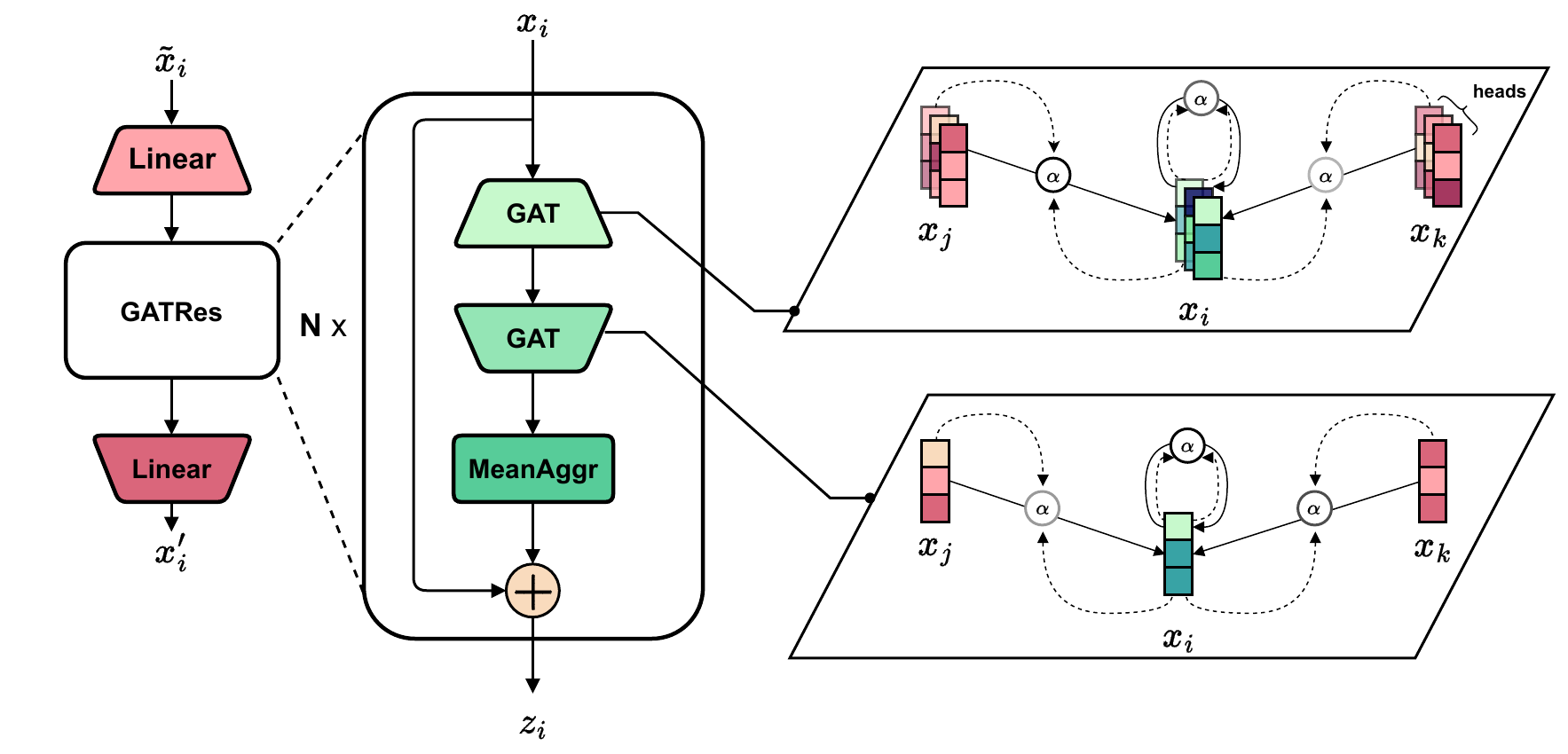}
    \caption{\textbf{\textit{GATRes} architecture.} The left image indicates the overall architecture consists of two linear layers interleaving with \textit{GATRes} blocks. The middle figure illustrates the abstract view in each block. The right-side ones explain the message aggregation mechanism between neighbor nodes. }
    \label{img:architect}
\end{figure} 

As in the middle image in Figure \ref{img:architect}, we feed the intermediate input $x_i$ to two GAT layers sequentially. For a target node $i$, the inner $GAT: \mathbb{R}^{N \times d_{in}} \rightarrow \mathbb{R}^{N \times Hd_{out}}$ devises multi-head views and weighs them among its 1-hop neighbors. In other words, it additionally enriches the diversity of multi-head views using the message aggregation from surrounding nodes.

Then, the outer $GAT: \mathbb{R}^{N \times Hd_{out}} \rightarrow \mathbb{R}^{N \times d_{out}}$ creates a bottleneck in the feature dimension and, again, reweights the target representation considering the ones of its neighbors. Note that the second GAT has exactly one head to transform all previous heads into a consistent view. We call this process a squeezing technique. Since most initial features are noise or zeros, squeezing can reduce the sparsity duplication in feature space caused by head concatenation and, therefore, benefits second attention among nodal pairs. Moreover, we consider the distribution of pressure values in the neighborhood, so we empirically apply a mean aggregator to the current representations \citep{xu2019gin}. Afterward, we use a non-parametric residual connection with the intermediate input $x_i$ that allows depth extension and diminishes the overfitting problem \citep{he2016deep}. 

As in Figure \ref{img:architect}, the overall structure is a stack of numerous \textit{GATRes} blocks. As each block considers 1-hop neighborhoods, stacking multi-blocks allows message propagation to faraway neighbors in the graph. Before message propagation layers, we employ a shared-weight linear transformation to project the masked input nodal features to higher-dimensional space. The details of masked inputs will be explained in the following subsection. We refer to the first linear layer as the steaming layer, which is well-known in computer vision tasks \citep{dosovitskiy2021ViT, tan2019efficientnet}. After message propagation, the final linear layer acts as a decoder to project higher-dimensional representations back to the original dimension (i.e., $d_{node}=1$). The end-to-end model will then output an immediate snapshot in which all pressure values at any junctions are recovered.

%\quickwordcount{model_architecture}

\subsection{Model training} % Huy
\label{sec:model_training}

This section introduces our solution to leverage \textit{GATRes} to solve the pressure estimation task. We first describe the general training scheme applied to any GNN model. Then, we provide an approach to test the trained model on time-relevant data. We also interpret why the proposed solution satisfies the criteria found in Section \ref{sec:related_work}.

\subsubsection{Training details}
Pressure estimation is a semi-supervised node-level task. Concretely, it aims to predict missing node features in the entire graph, given limited known nodal information and graph-related properties. Due to the lack of historical data and sensor scarcity, we leverage nodal features $\mathcal{X}$ created in our data generation to form the training set.

To begin with, we sample a binary mask vector $m = \{m_1,m_2,..., m_N\}$ where each $m_i \in \{0, 1\}$. We then construct a feature subset of the masked node $\tilde{\mathcal{X}} \subset \mathcal{X}$ in which its element $\tilde{x}_i$ is denoted as:
\begin{equation} \label{eqn:mask}
    \tilde{x}_{i}=\begin{cases}
                        0 & m_i=1 \\
                        x_i & m_i=0
                  \end{cases}
\end{equation}
There exist various masking strategies, such as learnable [MASK] tokens, feature permutation, arbitrary vector substitution, and mixup nodal features, which could be helpful for future work. In this work, we opt for a simple masking approach: replacing the node features with zeros in the masked positions to create the masked $\tilde{\mathcal{X}}$ (Equation \ref{eqn:mask}). We then formalize the pressure estimation as follows:
\begin{equation}
    \mathcal{X}' = f_{GNN}(\tilde{\mathcal{X}}, \mathcal{A}, \mathcal{E}; \Theta)
\end{equation}
where $f_{GNN}: \mathbb{R}^{N\times d_{node}} \times \mathbb{R}^{N\times N} \times \mathbb{R}^{M\times d_{edge}} \mapsto \mathbb{R}^{N\times d_{node}}$ is a generic GNN function that takes partial-observable feature matrix $\tilde{\mathcal{X}}$, the topology $\mathcal{A}$, and edge attributes $\mathcal{E}$ as inputs and yields reconstructed features $\mathcal{X}'$ characterized by model weights $\Theta$. The key idea is to find the optimal weights that satisfy the minimum error between predicted and ground truth nodal features. Mathematically, the objective is formalized as follows:
\begin{equation}
    \Theta^* = \operatorname*{argmin}_\Theta \mathcal{L}(\mathcal{X}', \mathcal{X})
\end{equation}
Empirically, we use the mean square error (MSE) as a default loss function $\mathcal{L}$ for \textit{GATRes} because it yields the best result in our tests. In addition, inspired by BERT \citep{devlin2018bert}, the loss is computed on masked positions.
%only to prevent the model from memorizing values when its size is enormous enough (i.e. data leakage problem).
After computation, model weights $\Theta$ are updated by the partial derivatives w.r.t the computed loss. For detail, we refer to gradient descent optimization techniques \citep{ruder2016sgd, kingma2014adam}
. The training progress is then iterated with different masked features $\tilde{\mathcal{X}}$ derived from the original features $\mathcal{X}$ until the model convergence.
\begin{figure}[h]
    \centering
    \noindent\includegraphics[width=0.75\textwidth]{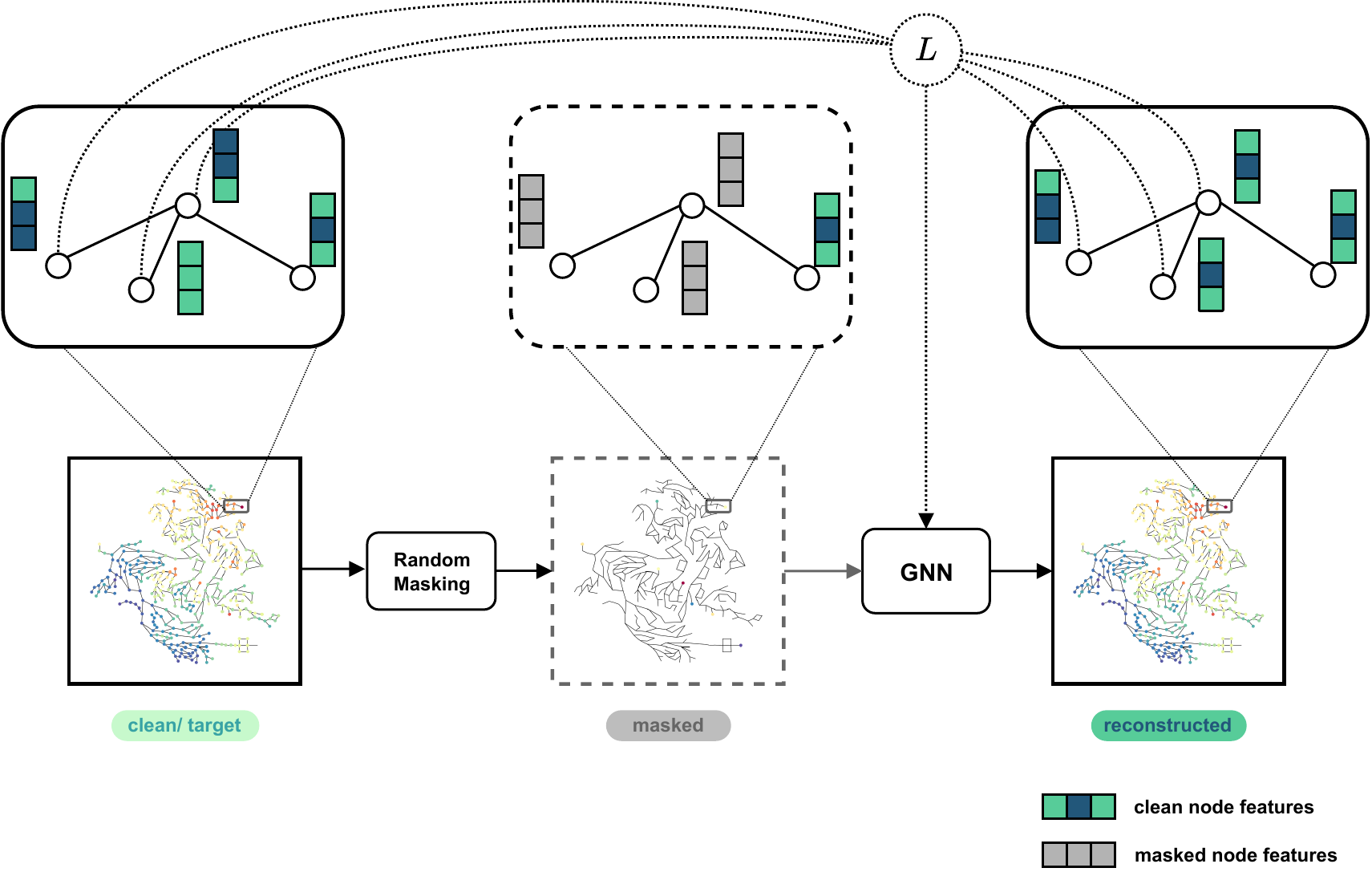}
    \caption{\textbf{Graph Neural Network training scheme.} Given clean snapshots, we mask out a significant number (95\%) of node features. The remaining data is then sent into a GNN playing as an autoencoder to rebuild missing values with regard to graph properties (such as topology and edge attributes). GNN weights are updated using the loss derived by the predicted and ground truth values at masked places.}
    \label{img:model_training}
\end{figure} 

\subsubsection{Testing details}
In testing, the test graphs can be represented as $\mathcal{G}_{test}=\{\tilde{\mathcal{X}}_{test}, \mathcal{A}_{test}, \mathcal{E}_{test}\}$. By default, topology and edge attributes are retained as in training while $\tilde{\mathcal{X}}_{test}$ is varied and its data distribution is undetermined. For the generalization problem, $\mathcal{G}_{test}$ can differ from the training graphs in any property. In other words, the model should be able to estimate pressure on an unseen topology and an unknown data distribution.

With the time involved, the test graph at a particular time $t$ is denoted as $\mathcal{G}^t_{test}$. As the designed model takes only one snapshot as input, we feed temporal $\mathcal{G}^t_{test}$ into \textit{GATRes} sequentially and individually. As a result, the reconstructed outcome does not affect the consecutive reconstructions during inference phase. In other words, this simple strategy isolates the model decision from the tendency to time-related data rarely found in training. 

Developing data generation concerning time data and a temporal GNN is also possible, but it exponentially raises the cost of complexity and computation. Thus, we encourage further work to explore temporal models, timed data generation, and the trade-off between efficiency and performance in the future. 
%For instance, a temporal data generator needs to manage the time series of all nodes in the entire graph. It is no guarantee these signals represent individual junction behavior and keep consistency over the timeline, especially in a large-scale matter.
In contrast, snapshot-based generators and GNNs ensure efficiency that satisfies practical needs (e.g. inference time).

%the testing data is delivered by the same mathematical simulation but controlled by diverse recorded demand patterns.

\subsubsection{Criteria satisfaction}
\label{sec:crit_satis}
Figure \ref{img:model_training} illustrates the training scheme leveraging the synthetic dataset to solve the pressure estimation task on practical data. We remark that it satisfies the predefined criteria: $\boldsymbol{(\mathbb{C}1)}$ by \textit{GATRes} being a spatial-based GNN approach that has topology awareness in its decision, $\boldsymbol{(\mathbb{C}2)}$ by random masking that dynamically changes sensor positions and myriad contextual snapshots from our data generation tool, and $\boldsymbol{(\mathbb{C}3)}$ by effectively evaluating the model on unseen time-relevant data with respect to uncertainty conditions.

%\quickwordcount{model_training} 

\section{Experiment settings}
\label{sec:exp_settings}

\subsection{Datasets} % ANDRES

\label{sec:datasets}

\begin{figure}[h]
    \centering
    \noindent\includegraphics[width=0.9\textwidth]{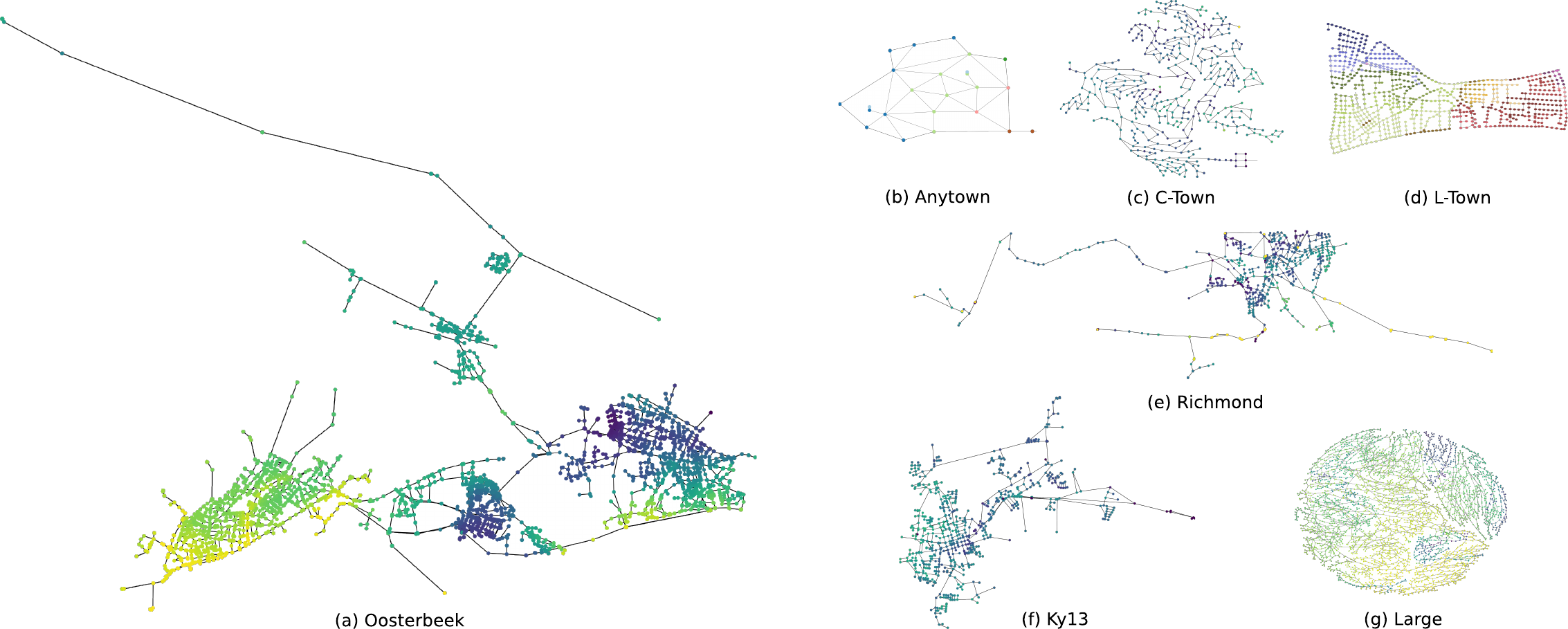}
    \caption{Water Distribution Networks used in this study.}
    \label{fig:WDNs}
\end{figure}

The main use case in this study was performed using a private large-scale WDN in The Netherlands in the area of Oosterbeek. The network comprises 5855 junctions and 6188 pipes. Figure \ref{fig:WDNs}(a) shows the topology and the pressures at the nodes from some random snapshot of Oosterbeek WDN. 

% \begin{figure}[h]
%     \centering
%     \noindent\includegraphics[width=0.75\textwidth]{../figures/oosterbeek.pdf}
%     \caption{Oosterbeek Water Distribution Network}
%     \label{fig:oosterbeek}
% \end{figure}	

We also used four publicly available WDNs benchmarks, namely Anytown \citep{walski1987battle}, C-Town \citep{ostfeld2012battle}, L-Town \citep{Vrachimis2022}, and Richmond \citep{van2001methodology} to provide a baseline for evaluation and reproducibility of our work . Finally, in the experiments related to model generalization, we used two additional public datasets, Ky13 \citep{hernadez2016water} and an anonymized WDN called ``Large'' \citep{sitzenfrei2023dual}. The WDNs used in this study vary is size and structure, ranging from small and medium size to large-scale networks like ``Large'' and Oosterbeek, as can be seen in Figure \ref{fig:WDNs}. Table \ref{tab:wdns_properties} shows the main characteristics of each network.

\begin{table}[h]
    \caption{Properties of WDNs used in this study.}
    \centering
    \label{tab:wdns_properties}
    \begin{tabular}{l c c c c c c c c}
        \hline
        WDNs: & Oosterbeek & Anytown & C-Town & L-Town & Richmond & Ky13 & Large  \\
        \hline
        $junctions$  & 5855 & 22 & 388 & 785 & 865 & 775 & 3557 \\
        $pipes$  & 6188 & 41 & 429 & 909 & 949 & 915 & 4021 \\
        \hline
        %\multicolumn{2}{l}{$^{a}$Footnote text here.}
    \end{tabular}
\end{table}

\subsection{Baseline models settings} %Huy

Generally, two goals dictate the baseline selection. Section \ref{sec:related_work} mentions the first goal: to try out popular GNN architectures on a node-level regression problem. The aim is to achieve acceptable errors when the models are tested on data points that are from an unknown ``nature'' distribution. The second goal is to explore existing frameworks, from synthesized data querying and model training to evaluation phases. We aim to establish a reliable benchmarking framework for pressure estimation tasks or problems related to water distribution networks. In other words, the model that performs better in our tests should be more useful in practical applications.

For this purpose, we compare our \textit{GATRes} architecture to popular GNNs, including \textit{GCNii} \citep{chen2020gcn2} and \textit{GAT} \citep{velickovic2018gat}. In addition to this, \textit{GraphConWat} (GCW) \citep{hajgato2021reconstructing} and \textit{mGCN} \citep{ashraf2023spatial}, which are dominant approaches in solving pressure estimation using GNN with sparse information, are also considered in our comparison. Table \ref{tab:baseline_parameters} summarizes the model settings. 
It is worth noting that we uniquely employ the best settings in each model across all considered WDNs to guarantee $\boldsymbol{(\mathbb{C}1)}$ satisfaction. 
%TODO: ADD table

\begin{table}[h]
\caption{\textbf{Baseline settings.} }
\centering
\label{tab:baseline_parameters}
\resizebox{\columnwidth}{!}{%p{0.15\linewidth}
    \begin{tabular}{@{\quad}l *{8}{c} @{\quad}}
    \toprule
    
            & GCNii & GAT &  GCW & \pbox{0.11\linewidth}{GCW\\tuned\\(ours)}  & mGCN & \pbox{0.11\linewidth}{GATRes\\small\\(ours)} & \pbox{0.11\linewidth}{GATRes\\large\\(ours)} \\\midrule
            
    \#blocks& 64 & 10 &  4 & 4 & 45 & 15 & 25\\
    \#hidden.channels& 32 & \{32,64\} &  \{120,60,30\} & 32 & \{98,196\} & \{32,64\} & \{128,256\} \\
    Coefficient K& -  & - & \{240,120,20,1\} & \{24,12,10,1\}  & - & - & -\\
    Loss& MSE & MSE & MSE &  MSE & MAE & MSE & MSE\\
    Learning Rate& 3e-4 & 3e-4 &  3e-4 & 3e-4 & 1e-5 & 5e-4 & 5e-4\\
    Weight Decay& 1e-6 & 1e-6 &  1e-6 & 1e-6 & 0 & 1e-6  & 1e-6 \\
    Edge Attribute& binary & binary & binary &binary & pipe.len, pipe.dia$^{a}$  & binary & binary\\
    Norm Type& znorm & znorm &  minmax & znorm & minmax & znorm & znorm
    \\\bottomrule

    %Model & \#blocks & \#channels & Loss & Edge.Attr &Lr &Weight.Decay &Norm.Type\\\midrule
    %GCNii & 100 & 100 & 100 & 100 & 100 & 100 & 100 \\\midrule
    \multicolumn{9}{l}{$^{a}$ Pipe lengths and pipe diameters. They are static parameters gathered from the corresponding Water Distribution Network.}

    \end{tabular}
    
}
\end{table}

% \begin{table}[h]
% \caption{\textbf{Baseline settings.} }
% \centering
% \label{tab:baseline_parameters}
% \resizebox{0.98\columnwidth}{!}{%p{0.15\linewidth}
%     \begin{tabular}{@{\quad}l *{9}{c} @{\quad}}
%     \toprule
    
%             & GCNii & GAT & GIN & GCW & \pbox{0.11\linewidth}{GCW\\tuned\\(ours)}  & mGCN & \pbox{0.11\linewidth}{GATRes\\small\\(ours)} & \pbox{0.11\linewidth}{GATRes\\large\\(ours)} \\\midrule
            
%     \#blocks& 64 & 10 & 15 & 4 & 4 & 45 & 15 & 25\\
%     \#hidden.channels& 32 & \{32,64\} & \{32,16\} & \{120,60,30\} & 32 & \{98,196\} & \{32,64\} & \{128,256\} \\
%     Coefficient K& -  & - & - & \{240,120,20,1\} & \{24,12,10,1\}  & - & - & -\\
%     Loss& MSE & MSE & MSE & MSE & MSE & MAE & MSE & MSE\\
%     Learning Rate& 3e-4 & 3e-4 & 5e-4 & 3e-4 & 3e-4 & 1e-5 & 5e-4 & 5e-4\\
%     Weight Decay& 1e-6 & 1e-6 & 1e-6 & 1e-6 & 1e-6 & 0 & 1e-6  & 1e-6 \\
%     Edge Attribute& binary & binary & binary & binary &binary & pipe.len, pipe.dia$^{a}$  & binary & binary\\
%     Norm Type& znorm & znorm & znorm & minmax & znorm & minmax & znorm & znorm
%     \\\bottomrule

%     %Model & \#blocks & \#channels & Loss & Edge.Attr &Lr &Weight.Decay &Norm.Type\\\midrule
%     %GCNii & 100 & 100 & 100 & 100 & 100 & 100 & 100 \\\midrule
%     \multicolumn{9}{l}{$^{a}$ Pipe lengths and pipe diameters. They are static parameters gathered from the corresponding Water Distribution Network.}

%     \end{tabular}
    
% }
% \end{table}

\textit{GATRes-small} with hyperparameters is the optimal version after the optimization process, which will be carefully explained in the latter section. To study the impact of the model size, we also introduce \textit{GATRes-large} scaling close to \textit{mGCN} in terms of the number of parameters.

\textit{GAT} remained at a shallow depth to prevent the oversmoothing problem \citep{chen2020oversmoothing}. Precisely, neighbor features encoded by a too-deep GNN converged to indistinguishable embeddings that harm the model performance. Empirically, we balance the trade-off between performance and efficiency for each model to select the appropriate hyperparameters.

In \textit{GraphConvWat} models, we detached binary masks from the input features as these masks did not improve the model performance, which aligns with the findings in \citep{ashraf2023spatial}. Furthermore, \textit{ GraphConvWat tuned} is a lightweight version in which the degrees of the Chebyshev polynomial $K_{i}$ are set to smaller values to reduce complexity and work surprisingly well in our experiments.

Training \textit{mGCN} slightly diverged from its original work in sensor positions. Concretely, \citep{ashraf2023spatial} trained \textit{mGCN} using fixed sensors with extensive historical data. However, as we explained in Section \ref{sec:model_training}, this data was inaccessible throughout training. Therefore, we fed different random masks into the model in each epoch. Considering a synthetic dataset, the model had an opportunity to capture meaningful patterns in various sensor positions.

%\quickwordcount{baseline_models} 

\subsection{Evaluation metrics}% Andres

The most common evaluation metrics used for assessing the performance of regression models are Root Mean Square Error (RMSE), Mean Absolute Error (MAE) and Mean Absolute Percentage Error (MAPE) \citep{zhao2019t, derrow2021eta, jiang2022graph}. We consider important to use evaluation metrics which are common not only in the Machine Learning domain, but also in hydrologic sciences. Following the insights from \citep{Legates1999evaluating}, the models should be evaluated using both, relative and absolute error metrics. Thus, our model is evaluated using MAE and MAPE. Additionally, we included the Nash and Sutcliffe Coefficient of Efficiency (NSE), widely used to evaluate the performance of hydrologic models \citep{Legates1999evaluating}. Finally, we used an accuracy metric defined as the ratio of positive predictions over the total number of predicted values. The positive predictions are those that deviates at most a certain threshold $(\delta_{\text{thresh}})$ from the true value. Thus, the evaluation metrics used in this work are defined as follows:

\begin{equation}
	\text{MAE} = \frac{1}{N} \sum_{i=1}^{N} | y_i - \hat{y}_i |
\end{equation}

\begin{equation}
	\text{MAPE} = \frac{1}{N} \sum_{i=1}^{N} \frac {| y_i - \hat{y}_i |}{y_i}
\end{equation}

\begin{equation}
	\text{NSE} = 1 - \frac {\sum_{i=1}^{N}  ({y_i - \hat{y}_i})^2}{\sum_{i=1}^{N}  ({y_i - \overline{y}})^2}
\end{equation}

\begin{equation}
    acc(@\delta_{\text{thresh}}) = \frac{1}{N} \sum_{i=1}^{N} {positive_i}~; ~~~ 
    positive =
            \begin{cases}
                1, & \text{if}~ | y_i - \hat{y}_i | \leq \delta_{\text{thresh}} * y_i \\ 
                0, & \text{otherwise}
            \end{cases}   
\end{equation}

where $\bm{y}$ denotes the true values, $\bm{\hat{y}}$ denotes the predicted values, $\bm{\overline{y}}$ is the mean of the true values, and $\bm{N}$ is the number of values to predict.

\section{Experiments} %Experimental results
\label{sec:experiments}

\subsection{Baseline comparison on Oosterbeek WDN}% Huy
\label{sec:oos_comparison}

%\subsubsection{Experimental settings} 

In this experiment, we investigated the proposed model performance against GNN variants on a large-scale WDN benchmark called Oosterbeek. Specifically, given the topology and hydraulic-related parameters, our dataset generation provided 10,000 synthetic snapshots divided into 6000, 2000, and 2000 for training, validation, and testing sets, respectively. However, as we discussed in Section \ref{sec:phenomenon}, these synthetic sets might not reflect real-world scenarios. Therefore, we merely used them to keep track of model learning during the training process.

Alternatively, we performed the comparison on the Oosterbeek dataset recorded every five minutes for 24 hours. We relied on mathematical simulation to produce reproducible results that resembled real-world conditions. The topology and predefined parameters set by hydraulic experts under a calibration process made them valid for our analysis. As a result, we considered the simulated outcomes of time-relevant data as ground truths. However, it was essential to acknowledge that specific hydraulic parameters, such as customer demands and pipe roughness, remained undetermined due to the dimensional explosion in parameter space, leading to noticeable errors in practical scenarios \citep{zhou2023testing}. To replicate this uncertainty, we utilized a distortion approach on junction demands during the testing phase inspired by \citep{zhou2023testing, mucke2023sametestway}. 
We then listed two testing strategies in detail as follows.

\begin{enumerate}
    \setlength\itemsep{1em}
    \item[] \textbf{Clean test.} We assumed there was no uncertainty in this test so that baseline models could observe clean, calibrated simulation pressures. Noticeably, because they all considered snapshots as individual samples, we can sample random masks indicating the visible virtual sensors per snapshot in every run. Running 100 times with diverse measurement locations would show the model capability in perfect condition.
    
    \item[] \textbf{Noisy test.} Following \citep{zhou2023testing}, we injected Gaussian noise into junction demands before the simulation was processed and then paired each outcome snapshot with a random mask for a test case. We set a tougher noise that went beyond the original tests. The new noisy test involved the mean and standard deviation of 10\% and 100\% of the initial demands, respectively. We ran 100 test cases and reported statistical findings.
    
\end{enumerate}

\begin{table}[h]
\caption{\textbf{Model comparison in the clean test performed on 24-hour Oosterbeek WDN at 95\% masking rate.} }
\centering
\label{tab:oos_comparison}
\resizebox{0.98\columnwidth}{!}{%
\begin{tabular}{@{\quad}l c c c c c c@{\quad}}
\toprule
%https://wandb.ai/jcechan/model_evaluation_details?workspace=user-jcechan
%_test100_
Model & \pbox{0.1\linewidth}{\#Milion Params}($\downarrow$)&  MAE($\downarrow$) & \pbox{1.3cm}{MAPE}($\downarrow$)&NSE($\uparrow$) & Acc(@0.1)($\uparrow$)\\
\midrule
 GCNii \citep{chen2020gcn2}&0.65&
 6.357\scriptsize{$\pm$0.0197}&
 0.2147\scriptsize{$\pm$0.0008}&
 -0.0137\scriptsize{$\pm$0.0061}&
 38.48\scriptsize{$\pm$0.1351}\\\midrule
 
 GAT \citep{velickovic2018gat}&0.35&
 3.726\scriptsize{$\pm$0.0120}&
 0.1287\scriptsize{$\pm$0.0008}&
 0.3276\scriptsize{$\pm$0.0037}&
73.52\scriptsize{$\pm$0.0900}\\\midrule

 GraphConvWat \citep{hajgato2021reconstructing}&0.92&
 3.067\scriptsize{$\pm$0.0077}&
 0.1160\scriptsize{$\pm$0.0004}&
 0.6938\scriptsize{$\pm$0.0020}&
 69.92\scriptsize{$\pm$0.1205}\\
 
 GraphConvWat-tuned&\textbf{0.23}&
 2.293\scriptsize{$\pm$0.0087}&
 0.0821\scriptsize{$\pm$0.0005}&
 0.7518\scriptsize{$\pm$0.0024}&
 83.03\scriptsize{$\pm$0.1025}\\\midrule
 
 mGCN \citep{ashraf2023spatial}&2.48&
 2.111\scriptsize{$\pm$0.0085}&
 0.0806\scriptsize{$\pm$0.0003}&
 0.7100\scriptsize{$\pm$0.0030}&
 84.05\scriptsize{$\pm$0.0693}\\\midrule
 
 GATRes-small (ours)&0.66&
 \textbf{1.937}\scriptsize{$\pm$0.0074}&
 \textbf{0.0703}\scriptsize{$\pm$0.0005}&
 0.7773\scriptsize{$\pm$0.0025}&
 \textbf{87.48}\scriptsize{$\pm$0.0761}\\
 
 GATRes-large (ours)&1.67&
 2.020\scriptsize{$\pm$0.0132}&
 0.0711\scriptsize{$\pm$0.0003}&
 \textbf{0.7864}\scriptsize{$\pm$0.0031}&
 84.33\scriptsize{$\pm$0.1347}\\\bottomrule

\end{tabular}
}
\end{table}
\begin{table}[h]
\caption{\textbf{Model comparison in the noisy test performed on 24-hour Oosterbeek WDN at 95\% masking rate.}}
\centering
\label{tab:oos_comparison_noisy}
\resizebox{0.98\columnwidth}{!}{%
\begin{tabular}{@{\quad}l c c c c c c@{\quad}}
\toprule
%https://wandb.ai/jcechan/model_evaluation_details?workspace=user-jcechan
%_test100noisy11new_
Model & \pbox{0.1\linewidth}{\#Milion Params}($\downarrow$)&  MAE($\downarrow$) & \pbox{1.3cm}{MAPE}($\downarrow$)&NSE($\uparrow$) & Acc(@0.1)($\uparrow$)\\
\midrule
 GCNii \citep{chen2020gcn2}&0.65&
 6.696\scriptsize{$\pm$0.0838}&  
 0.2484\scriptsize{$\pm$0.0552}&  
 -0.1064\scriptsize{$\pm$0.0266}&
 36.02\scriptsize{$\pm$0.4684}
  \\\midrule
 
 GAT \citep{velickovic2018gat}&0.35&
 4.397\scriptsize{$\pm$0.3052}&
 0.2112\scriptsize{$\pm$0.0767}& 
 0.1490\scriptsize{$\pm$0.1153}&
 66.98\scriptsize{$\pm$1.6290}
 
 \\\midrule
 
 GraphConvWat \citep{hajgato2021reconstructing}&0.92& 
 3.611\scriptsize{$\pm$0.1234}&
 0.1551\scriptsize{$\pm$0.0376}&
 0.5877\scriptsize{$\pm$0.0370}&
 62.99\scriptsize{$\pm$1.1600}
 
 \\
 
 GraphConvWat-tuned&\textbf{0.23}&
 2.347\scriptsize{$\pm$0.0252}&
 0.0963\scriptsize{$\pm$0.0363} &  
 0.749\scriptsize{$\pm$0.0086}&
 81.09\scriptsize{$\pm$0.3877}

 \\\midrule
 
 mGCN \citep{ashraf2023spatial}&2.48&
 2.188\scriptsize{$\pm$0.0558}&   
 0.0948\scriptsize{$\pm$0.0155}&  
 0.6993\scriptsize{$\pm$0.0213}&
 82.83\scriptsize{$\pm$0.4199}

 \\\midrule
 
 GATRes-small (ours)&0.66&
 \textbf{1.964}\scriptsize{$\pm$0.0301}&
 0.0802\scriptsize{$\pm$0.0458}&  
 \textbf{0.778}\scriptsize{$\pm$0.0113}&
 \textbf{86.56}\scriptsize{$\pm$0.2826}
 \\
 
 GATRes-large (ours)&1.67&
 2.115\scriptsize{$\pm$0.0503}& 
 \textbf{0.0799}\scriptsize{$\pm$0.0207}&  
 0.7417\scriptsize{$\pm$0.0140}&
 83.43\scriptsize{$\pm$0.5044}
 
 \\\bottomrule

\end{tabular}
}
\end{table}

Regarding the experimental setting, we trained all models in 500 epochs with a batch size of 8 for a fair comparison among the baseline models. Early Stopping was applied to suppress training if the validation error had no improvements in 100 steps. We used Adam optimizer \citep{kingma2014adam} and set the default masking rate at 95\%, leaving only 5\% of nodes unmasked. 

For evaluation, we tested the baseline models on the 24-hour Oosterbeek, repeating the process a hundred times. The mean and standard deviation of the results are presented in Table \ref{tab:oos_comparison}. Unless otherwise specified, the default is a clean test in our experiments. The results of the noisy test are given in Table \ref{tab:oos_comparison_noisy}.

Table \ref{tab:oos_comparison} shows that \textit{GATRes-small} achieved accurate junction pressure reconstruction with an average relative error of 7\% and an absolute error of 1.93 water column meters, even with a sparse masking ratio of 95\%. Notably, the testing data were time-sensitive and originated from an unfamiliar distribution our models were not exposed to during training. The good results on snapshot-based models suggest that in case temporal data is not available, snapshot-based models seem to be good alternatives.
\begin{figure}[h]
    \centering
    \noindent\includegraphics[width=0.75\textwidth]{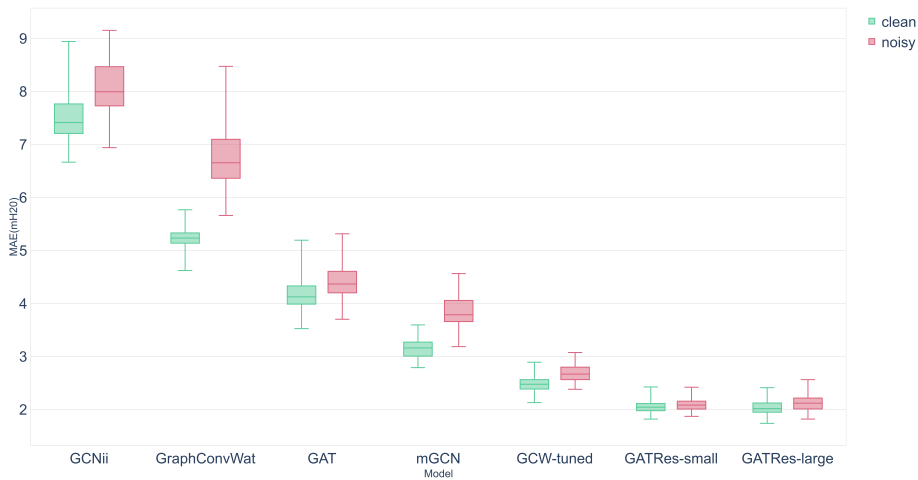}
    \caption{\textbf{Baseline Mean Absolute Errors measured for a single snapshot under both clean and noisy conditions.} 
    }
    \label{img:oos_testing}
    %https://wandb.ai/jcechan/model_evaluation_details_single?workspace=user-jcechan
\end{figure} 
%%move appendix to here
%%%%%%%%%%%%%%%%%%%%%%%%%%%%%%%%%%%%%%%%%%%%%%%%%%%%%%%%%%%%%%%

%$\section{GatRes performance on simulation-generated data}

%\section{Further experiments on the Oosterbeek WDN}
%\subsection{The efficiency comparison}
%We also measured the throughput, which indicates how many snapshots could be processed in a second. We employed a Nvidia RTX 3060 Laptop GPU during this comparison and reported the findings in Table \ref{tab:throughput}.

In addition, we assessed the efficiency of baseline models in the Oosterbeek experiment using an Nvidia RTX 3060 Laptop GPU for inference only. The results are presented in Table \ref{tab:throughput}. 

In this evaluation, we measured throughput, which counts the number of processed snapshots in a second. This metric can demonstrate the efficiency of baselines in terms of large-scale matter that demands continuous processing of massive data streams from sensors. 

Notably, lightweight models such as \textit{GAT} and \textit{GraphConvWat-tuned} achieved the highest throughput, with our \textit{GATRes-small} model ranking third. When we consider both efficiency and performance in Table \ref{tab:throughput}, \textit{GATRes-small} is a balanced option as this model delivers the best result while maintaining sufficient efficiency, a critical factor in saving computation resources and ensuring the sustainability of the environment.

% \begin{table}[h]
% \caption{\textbf{Throughput comparison in the noisy test performed on 24-hour Oosterbeek WDN at 95\% masking rate.} \textbf{Bold} and \underline{underline} indicates the best and second-best results, respectively.}
% \centering
% \label{tab:throughput}
% \resizebox{0.7\columnwidth}{!}{%
% \begin{tabular}{@{\quad}l c @{\quad}}
% \toprule
% %https://wandb.ai/jcechan/model_evaluation_details?workspace=user-jcechan
% %_test100noisy11new_
% Model & \parbox{0.4\linewidth}{\centering Throughput($\uparrow$)\\(Snapshots per second)}\\
% \midrule
%  GCNii \citep{chen2020gcn2}&646.23\\\midrule
 
%  GAT \citep{velickovic2018gat}&42.54\\\midrule
 
%  GraphConvWat \citep{hajgato2021reconstructing}&90.67\\
 
%  GraphConvWat-tuned&\textbf{2006.92}\\\midrule
 
%  mGCN \citep{ashraf2023spatial}&91.93\\\midrule
 
%  GATRes-small (ours)&\underline{748.16}\\
 
%  GATRes-large (ours)&128.15\\\bottomrule

% \end{tabular}
% }
% \end{table}

\begin{table}[h]
\caption{\textbf{Throughput comparison in the clean test performed on 24-hour Oosterbeek WDN at 95\% masking rate.} }
\centering
\label{tab:throughput}
\resizebox{0.7\columnwidth}{!}{%
\begin{tabular}{@{\quad}l c @{\quad}}
\toprule
%https://wandb.ai/jcechan/model_evaluation_throughput?workspace=user-jcechan
%_test100_
Model & \parbox{0.4\linewidth}{\centering Throughput($\uparrow$)\\(Snapshots per second)}\\
\midrule
 GCNii \citep{chen2020gcn2}&663.80\\\midrule
 
 GAT \citep{velickovic2018gat}&2320.37\\\midrule
 
 GraphConvWat \citep{hajgato2021reconstructing}&90.39\\
 
 GraphConvWat-tuned&2026.65\\\midrule
 
 mGCN \citep{ashraf2023spatial}&44.94\\\midrule
 
 GATRes-small (ours)&749.38\\
 
 GATRes-large (ours)&31.21\\\bottomrule

\end{tabular}
}
\end{table}

%%%%%%%%%%%%%%%%%%%%%%%%%%%%%%%%%%%%%%%%%%%%%%%%%%%%%%%%%%%%%%%
Our next focus was on analyzing the baseline robustness. Precisely, we assessed each baseline on clean and noisy tests using an individual snapshot with a hundred randomly initialized masks. Our primary objective was to measure the model's robustness in these contrasting scenarios. A superior model should exhibit minimal error discrepancy between them. As illustrated in Figure \ref{img:oos_testing}, both versions of \textit{GATRes} consistently maintained similar error levels even under conditions of high uncertainty. In contrast, other models exhibited a noticeable gap in their results when transitioning from clean to noisy environments.

\begin{figure}[h]
    \centering
    \includegraphics[width=0.75\textwidth]{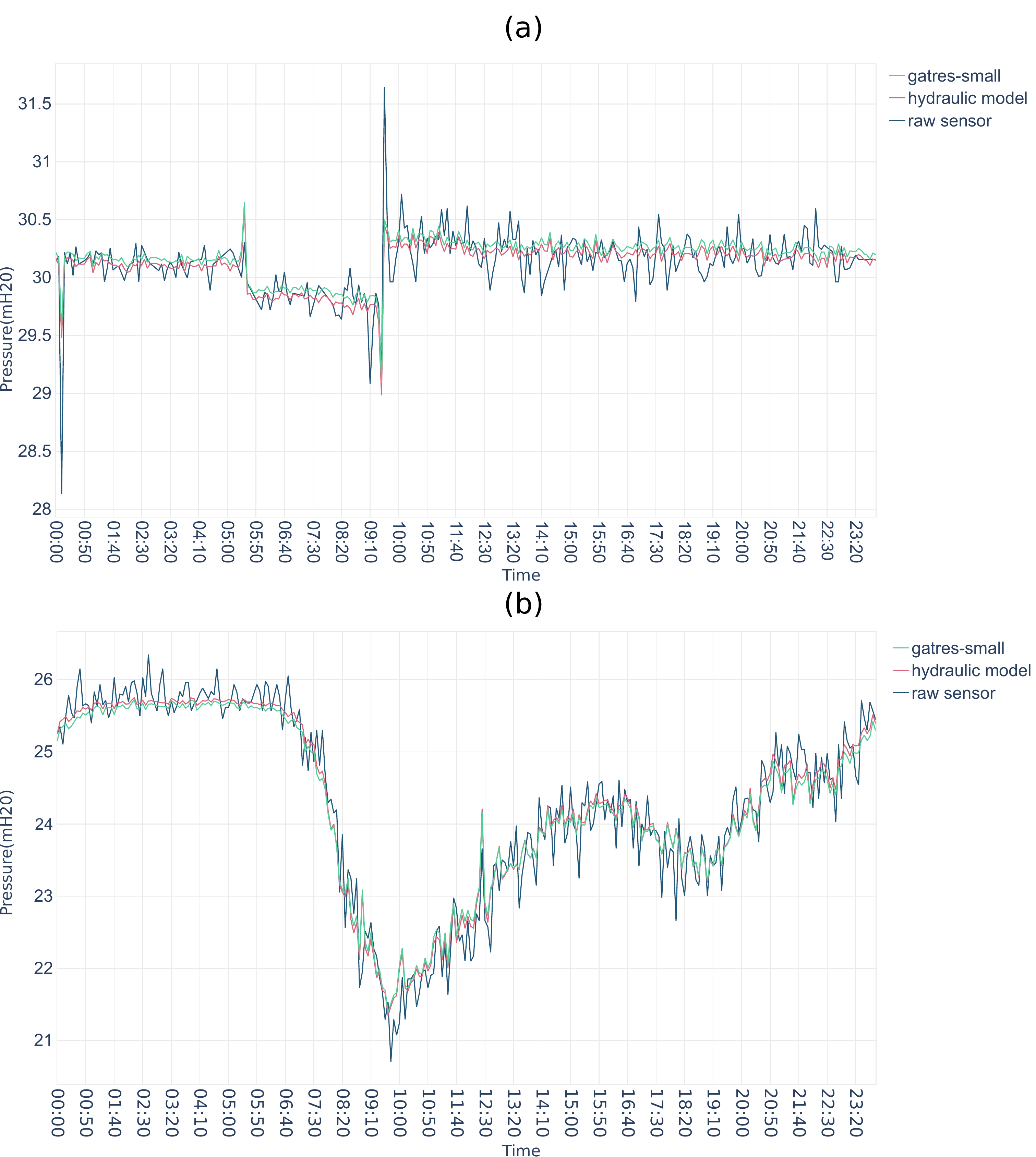}
    \caption{\textbf{Pressure values estimated on other actual sensors} }
    \label{img:timeseries}
\end{figure}

% \begin{figure}[h]
%     \centering
%     \noindent\includegraphics[width=0.9\textwidth]{figures/timeseries_GE-MOJA_0.05537865310907364.pdf}
%     \caption{\textbf{Estimated pressure of GATRes-small, hydraulic simulation, and a raw sensor meter A.} On a detailed scale, a slight discrepancy exists between our model(emerald) and the simulation(copper green), but its trend is bounded in the valid range of "real" measurements(dark blue).
%     }
%     \label{img:oss_timeseries_A}
% \end{figure} 

Finally, we conducted a detailed analysis of our top-performing model, \textit{GATRes-small}, based on the evaluations conducted earlier. In this analysis, we intentionally covered sensor locations to observe model inference on those nodes. Figure \ref{img:timeseries} illustrates time series data from the predictions of \textit{GATRes-small}, a well-calibrated simulation, and actual meter readings. As expected, \textit{GATRes} closely mirrors the behavior of the hydraulic simulation, demonstrating sufficient capability of a desired surrogate model. Although a slight difference exists between them, both time series were bounded in the range of actual measurements.

%\quickwordcount{oos_comparison} 

\subsection{Generalization}% ANDRES

	\label{sec:generalization}
		
	This set of experiments are the first attempts aimed to achieve generalization capabilities of our model. We want to evaluate whether training a model on different topologies simultaneously can equip the model with generalization capabilities. For this, we chose three different WDNs, which topologies vary in structure and size. 
	
	First, we trained \textit{GATRes-small} on L-Town, Ky13 and ``Large'' WDNs simultaneously. This model is named as Multi-Graph model. We wanted to evaluate the performance of this model on a fully unseen topology. Hence, we executed Zero-Shot inference (predict on a WDN topology not seen during training) using the Multi-Graph model on our main use case Oosterbeek WDN. This model was trained with the \textit{ReduceLROnPlateau} learning rate scheduler from Pytorch library. The scheduler reduces the learned rate if the model does not improve for a certain number of epochs. This model was trained with a batch size of 16 for 500 epochs. The initial learning rate was set to 5$e$-3 and it is reduced by a factor of $0.1$ if the validation loss does not improve for $30$ consecutive epochs.
	
	A second experiment is transfer learning, a technique motivated by the fact that humans use previously learned knowledge to solve new tasks faster or better \citep{pan2009survey}. Hence, the learned weights by a model trained on some particular network(s) can be transferred to train and improve the prediction capabilities of a model on a new, previously unseen WDN. We applied transfer learning and fine-tuning, which involves using the weights of a pre-trained model on a source dataset to initialize the weights of the new model that will be trained on the target dataset. 
	
	In our work, the pre-trained model is the Multi-Graph model, trained on L-Town, Ky13 and ``Large'', and the Fine-Tuned model has as the target the Oosterbeek WDN. Usually, during fine-tuning, the top layers of the pre-trained model are frozen and reused as feature extractors for the target data. We empirically found that unfreezing the entire model and retraining all layers produces better results. Thus, we initialized the weights of the target model with those of the pre-trained one. Then, we reduced the learning rate for training the target model to avoid completely changing the pre-trained weights during fine-tuning. The learning rate was reduced from 5$e$-3 in the source model to 1$e$-4 during fine-tuning. The target model was trained with a batch size of 8 for 200 epochs. The results of these experiments are shown in Table \ref{tab:generalization}. They reflect those of \citep{yosinski2014transferable} who also found that combining transfer learning with fine-tuning shows better performance than a model trained directly on a target dataset.

	\begin{table}[h]
	\caption{Generalization evaluation on 24-hour Oosterbeek WDN}
	\centering
	\label{tab:generalization}
	\resizebox{0.60\columnwidth}{!}{%
		\begin{tabular}{l c c c}
			\toprule
			 & MAE ($\downarrow$) & MAPE ($\downarrow$) & NSE ($\uparrow$) \\
			\midrule
			GATRes-small  & 1.9370 \small{$\pm$0.0074} & 0.0703 \small{$\pm$0.0005} & 0.7773 \small{$\pm$0.0025}  \\
			Multi-Graph (Zero-Shot) &  3.0597 \small{$\pm$0.0074} & 0.0998 \small{$\pm$0.0005} & 0.5700 \small{$\pm$0.0045} \\
			Fine-tuned  & \textbf{1.9097} \small{$\pm$0.0076} & \textbf{0.0695} \small{$\pm$0.0005} & \textbf{0.7980} \small{$\pm$0.0030} \\
			\bottomrule
			%\multicolumn{2}{l}{$^{a}$Footnote text here.}
		\end{tabular}}
	\end{table}	
	
	%The main goal of this experiment is to compare the performance of the models trained on a single topology against the performance of the one trained on multiple graphs simultaneously. The results show that the model trained on multiple topologies  

\subsection{The effect of masking ratios}% Huy

To explore the model capability, we investigated the \textit{GATRes-small} on myriad masking rates. Identical to the previous experiment,  the model was trained on the synthetic dataset generated from our algorithm and performed a clean test on the 24-hour data. Both were devised from the Oosterbeek WDN. In addition, each \textit{GATRes-small} corresponding to a specific mask rate was trained within 200 epochs with the default settings. For convenience, we replaced the model name with fixed masking rates in this experiment.

\begin{figure}[h]
\centering
\includegraphics[width=0.75\textwidth]{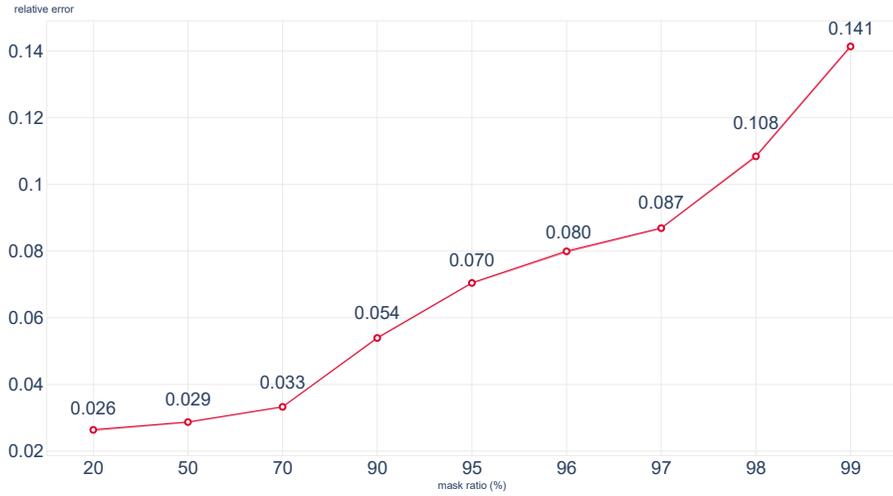}
\caption{\textbf{Relative errors (MAPE) for nodal pressure on different masking ratios(lower is better).} 
%Note that the same masking rates are conserved in both phases.
}
\label{img:diff_masks}
\end{figure}
Figure \ref{img:diff_masks} shows the influence of the masking ratio on the proposed model. Each ratio indicates a specific probability of missing nodal features (i.e., the pressure signals in a snapshot graph). Due to the sensor density being exceptionally sparse in real-world scenarios, the typical benchmark of 95\%, commonly found in previous studies, is deficient in reflecting this practical issue. Therefore, we report errors occurring in all cases with lower and more extreme ratios that exceed the standard. Additional metrics are found in Table \ref{tab:diff_mask}.

\begin{table}[h]
\caption{\textbf{Detailed performance of \textit{GATRes-small} on different masking ratios.} }
\centering
\label{tab:diff_mask}
    
\resizebox{0.6\columnwidth}{!}{%
\begin{tabular}{@{\quad}c c c c c@{\quad}}
\toprule
Mask ratio(\%) & MAE($\downarrow$) & MAPE($\downarrow$)&NSE($\uparrow$) & Acc(@0.1)($\uparrow$)\\\midrule

20 & 0.610 & 0.0265 & 0.9599 & 0.9760\\

50 & 0.798 & 0.0286 & 0.9372 & 0.9654\\

70 & 0.900 & 0.0331 & 0.9301 & 0.9586\\

90 & 1.457 & 0.0544 & 0.8603 & 0.9167\\

95 & 1.939 & 0.0703 & 0.7770 & 0.8746 \\

96 & 2.213 & 0.0800  & 0.7185  & 0.8467 \\

97 & 2.415 & 0.0867  & 0.7059  & 0.8148 \\

98 & 3.075 & 0.1091  & 0.5648  & 0.7465 \\

99 & 4.087 & 0.1414  & 0.3424  & 0.6396 \\\bottomrule
\end{tabular}
}
\end{table}

We conducted an additional investigation into the discrepancy between the masking ratios of train-test pairs. Our approach involved evaluating a trained model on the 24-hour Oosterbeek with various masking rates rather than just the specific rate initially trained. Through this exploration, we found that the best model of a specific testing masking rate was unnecessary to be trained on this rate. Table \ref{tab:mask_traintest} showed this phenomenon in extreme ratios. It could be seen that the model originally trained on 97\% could yield acceptable results on average. In addition, it surprisingly achieved the best results in extremely sparse test rates (i.e., $> 98\%$). This means that at most 3\% of the total nodes would be sufficient for a quality model to monitor the Oosterbeek WDN -- a large-scale network. Further analysis is highly recommended for WDN authorities to balance the trade-off between efficiency and measurement resources.

\begin{table}[h]
\caption{\textbf{Confusion matrix of relative mean errors (MAPE) between different train and test masking ratios(lower is better).} \textbf{Bold} and \underline{underline} are used to highlight the best and second-best results for a specific test mask, respectively.}
\centering
\label{tab:mask_traintest}
\begin{tabular}{l c c c c c}
\toprule
\multirow{2}{*}{\pbox{1.5cm}{test mask(\%)}}  &\multicolumn{5}{c}{train mask (\%)}\\ \cmidrule(lr){2-6} 
& 95 & 96 & 97 & 98 & 99\\\midrule

95 & \textbf{0.0702} & \underline{0.0723} & 0.0725 & 0.0772 & 0.0814\\\midrule

96 & \textbf{0.0766 }& 0.0797 & \underline{0.0784} & 0.0843 & 0.0882\\\midrule

97 & \textbf{0.0858} & 0.0901 & \underline{0.0870} & 0.0934 & 0.0970\\\midrule

98 & \underline{0.1031} & 0.1077 & \textbf{0.1018} & 0.1090 & 0.1109\\\midrule

99 & 0.1454 & 0.1494 & \textbf{0.1388} & \underline{0.1450} & 0.1414\\\bottomrule

\end{tabular}
\end{table}

\subsection{Baseline comparison on benchmark WDNs}% Andres

	In this set of experiments we compared the performance of \textit{GATRes-small} against two state-of-the-art baseline models, GraphConvWat \citep{hajgato2021reconstructing} and mGCN \citep{ashraf2023spatial}. We evaluated the three models on four benchmark WDNs: Anytown, C-Town, Richmond, and L-Town, described in Section \ref{sec:datasets}. The data used in these experiments were created following the method proposed by \citep{hajgato2020datagen} to facilitate comparability under the original conditions defined by previous works. We created 1,000 snapshots for Anytown, 10,000 snapshots for C-Town and L-Town, and 20,000 snapshots for Richmond. Then, the datasets were split into training, validation and test sets in a 6:2:2 ratio.	
	
	We used the same experimental settings as proposed in the baseline approaches to guarantee a fair comparison. Thus, in all experiments the models are trained for 2,000 epochs with early stopping, using the Adam gradient-based optimization algorithm \citep{kingma2014adam}. The GraphConvWat model training was stopped if the validation loss did not improve  for 50 consecutive epochs. In the case of mGCN, the training was stopped if no improvement is seen after 250 epochs. Likewise mGCN, our model training is stopped after 250 epochs if no improvement is observed. In all cases, it is considered an improvement when the validation loss decreases at least by \textit{1e-6}. 
	
	The evaluation of the models' performance on each WDN, with the exception of Anytown, was using data that included realistic demand patterns per node. In the case of C-Town and Richmond, the WDN snapshots for evaluation were created using a 24-hour demand pattern time series sampled at 5 minutes interval. L-Town evaluation snapshots were created using a 1-week demand pattern time series sampled at 5 minutes interval. Table \ref{tab:bsln_comp_time_inv} shows the results of the performance comparison of ten runs per WDN, and then the mean and standard deviation are reported. As can be seen from the table, our model \textit{GATRes-small} achieves the lowest MAE in all WDNs and the lowest MAPE in all networks but Richmond. Likewise, \textit{GATRes-small} achieves the higher NSE in all WDNs but Richmond.

	\begin{table}[h]
		\caption{Models performance comparison on 24-hour demand pattern time series data.}
		\centering
		\begin{tabular}{lrrrr}
			\toprule
			\multirow{2}{*}{{\small WDN}}
			& \multirow{2}{*}{{\small Metrics}}
			& \multicolumn{3}{c}{\small Models}
			\\
			\cmidrule(lr){3-5}
			
			& 
			& \multicolumn{1}{c}{\footnotesize GraphConvWat} 
			& \multicolumn{1}{c}{\footnotesize mGCN} 
			& \multicolumn{1}{c}{\footnotesize GATRes-small (ours)} 
			\\
			\hline
			
			\multirow{3}{*}{\footnotesize C-Town} 
			& \footnotesize MAE ($\downarrow$) 
			& 14.8860 \small{$\pm$0.1418} 
			&  19.9138 \small{$\pm$0.0948}
			& \textbf{09.4860} \small{$\pm$0.1822} 
			\\ 
			& \footnotesize MAPE ($\downarrow$) 
			& 0.1028 \small{$\pm$0.0009} 
			& 0.1318 \small{$\pm$0.0005} 
			& \textbf{0.0690} \small{$\pm$0.0010} 
			\\ 
			& \footnotesize NSE ($\uparrow$) 
			& 0.7870 \small{$\pm$0.0046} 
			& 0.6310 \small{$\pm$0.0030}
			& \textbf{0.8480} \small{$\pm$0.0075} 
			\\
			\cmidrule(lr){2-5}
			
			\multirow{3}{*}{\footnotesize Richmond} 
			& \footnotesize MAE ($\downarrow$) 
			& 4.3501 \small{$\pm$0.0170} 
 			& 2.9690 \small{$\pm$0.0283} 
 			& \textbf{2.8114} \small{$\pm$0.0899} 
			\\ 
			& \footnotesize MAPE ($\downarrow$) 
			& 0.0196 \small{$\pm$0.0001} 
			& \textbf{0.0128} \small{$\pm$0.0002}
			& 0.0133 \small{$\pm$0.0005} 
			\\ 
			& \footnotesize NSE ($\uparrow$) 
			& 0.9500 \small{$\pm$0.0000}  
			& \textbf{0.9630} \small{$\pm$0.0046}
			& 0.9390 \small{$\pm$0.0030}
			\\
			\cmidrule(lr){2-5}
			
			\multirow{3}{*}{\footnotesize L-Town} 
			& \footnotesize MAE ($\downarrow$) 
			& 3.4505 \small{$\pm$0.0129} 
			& 1.5928 \small{$\pm$0.0050} 
			& \textbf{0.9501} \small{$\pm$0.0086} 

			\\ 
			& \footnotesize MAPE ($\downarrow$) 
			& 0.0611 \small{$\pm$0.0002} 
			& 0.0305 \small{$\pm$0.0001}
			& \textbf{0.0157} \small{$\pm$0.0002}
			\\ 
			& \footnotesize NSE ($\uparrow$) 
			& 0.5040 \small{$\pm$0.0049} 
			& 0.8000 \small{$\pm$0.0000}
			& \textbf{0.9000} \small{$\pm$0.0000} 
			\\			
						
			\bottomrule
		\end{tabular}
	\label{tab:bsln_comp_time_inv}
	\end{table}
	
%	The evaluation data for Anytown were created following the same approach than \citep{hajgato2020datagen} because realistic demand patterns are not available for this network.
	
	One limitation of previous approaches is the evaluation of model performance on unrealistic data, i.e., an exact copy of the training data distribution (Section \ref{sec:phenomenon}). In previous approaches, the snapshots representing random WDN states used for training, validation and test are created by the same algorithm. Consequently, the distribution of the data used for testing is a fidelity copy of the data used for training. However, in practice, the distribution of the real data differs from the data used for training the reconstruction models, as explained in Section \ref{sec:phenomenon}. Therefore, it is important that the models adapt to circumvent such uncertainties. Previous approaches achieve impressive performance when tested on replicas of the training data (see Table \ref{tab:bsln_comp_hajgato_inv}), but the performance drop is evident when they are evaluated on a realistic scenario (see Table \ref{tab:bsln_comp_time_inv}).

	\begin{table}[ht]
		\caption{Models performance comparison on synthetic sampling-based snapshots following \citep{hajgato2020datagen} approach.}
		\centering
		\begin{tabular}{lrrrr}
			\toprule
			\multirow{2}{*}{{\small WDN}}
			& \multirow{2}{*}{{\small Metrics}}
			& \multicolumn{3}{c}{\small Models}
			\\
			\cmidrule(lr){3-5}
			
			& 
			& \multicolumn{1}{c}{\footnotesize GraphConvWat} 
			& \multicolumn{1}{c}{\footnotesize mGCN} 
			& \multicolumn{1}{c}{\footnotesize GATRes-small (ours)} 
			\\
			\hline
			
			\multirow{3}{*}{\footnotesize Anytown} 
			& \footnotesize MAE ($\downarrow$)
			& 5.1044 \small{$\pm$0.0714} 
			&  3.9460 \small{$\pm$0.0642} 
			& \textbf{3.9245} \small{$\pm$0.1056} 
			\\
			& \footnotesize MAPE ($\downarrow$)
			& 0.0654 \small{$\pm$0.0012} 
			& 0.0497 \small{$\pm$0.0009}
			& \textbf{0.0491} \small{$\pm$0.0012}
			\\ 
			& \footnotesize NSE ($\uparrow$)
			& 0.7440 \small{$\pm$0.0049} 
			& \textbf{0.8020} \small{$\pm$0.0075}
			& 0.7980 \small{$\pm$0.0189}
			\\			
			\cmidrule(lr){2-5}
			
			\multirow{3}{*}{\footnotesize C-Town} 
			& \footnotesize MAE ($\downarrow$)
			& 4.1619 \small{$\pm$0.0170} 
			& \textbf{1.6963} \small{$\pm$0.0133} 
			& 1.8928 \small{$\pm$0.0149} 
			\\ 
			& \footnotesize MAPE ($\downarrow$) 
			& 0.0354 \small{$\pm$0.0001} 
			& \textbf{0.0148} \small{$\pm$0.0001}
			& 0.0169 \small{$\pm$0.0001}
			\\ 
			& \footnotesize NSE ($\uparrow$) 
			& 0.9640 \small{$\pm$0.0049} 
			& \textbf{0.9900} \small{$\pm$0.0000}
			& \textbf{0.9900} \small{$\pm$0.0000}
			\\
			\cmidrule(lr){2-5}
			
			\multirow{3}{*}{\footnotesize Richmond} 
			& \footnotesize MAE ($\downarrow$) 
			& 2.3999 \small{$\pm$0.0069} 
			& \textbf{0.6363} \small{$\pm$0.0061} 
			& 1.5979 \small{$\pm$0.0106} 
			\\ 
			& \footnotesize MAPE ($\downarrow$) 
			& 0.0110 \small{$\pm$0.0000} 
			& \textbf{0.0029} \small{$\pm$0.0000} 
			& 0.0080 \small{$\pm$0.0001} 
			\\ 
			& \footnotesize NSE ($\uparrow$) 
			& 0.9805 \small{$\pm$0.0003}
			& \textbf{0.9900} \small{$\pm$0.0000}
			& 0.9750 \small{$\pm$0.0009}
			\\
			\cmidrule(lr){2-5}
			
			\multirow{3}{*}{\footnotesize L-Town} 
			& \footnotesize MAE ($\downarrow$) 
			& 1.2970 \small{$\pm$0.0036} 
			& \textbf{0.2441} \small{$\pm$0.0014} 
			& 0.4930 \small{$\pm$0.0028} 
			\\ 

			& \footnotesize MAPE ($\downarrow$) 
			& 0.0159 \small{$\pm$0.0000} 
			& \textbf{0.0030} \small{$\pm$0.0000} 
			& 0.0061 \small{$\pm$0.0000} 
			\\ 

			& \footnotesize NSE ($\uparrow$) 
			& 0.9700 \small{$\pm$0.0000}
			& \textbf{1.0000} \small{$\pm$0.0000}
			& \textbf{1.0000} \small{$\pm$0.0000} 
			\\			
			
			\bottomrule
		\end{tabular}
		\label{tab:bsln_comp_hajgato_inv}
	\end{table}
	
\begin{figure}[ht]
		\centering
		\noindent\includegraphics[width=0.95\textwidth]{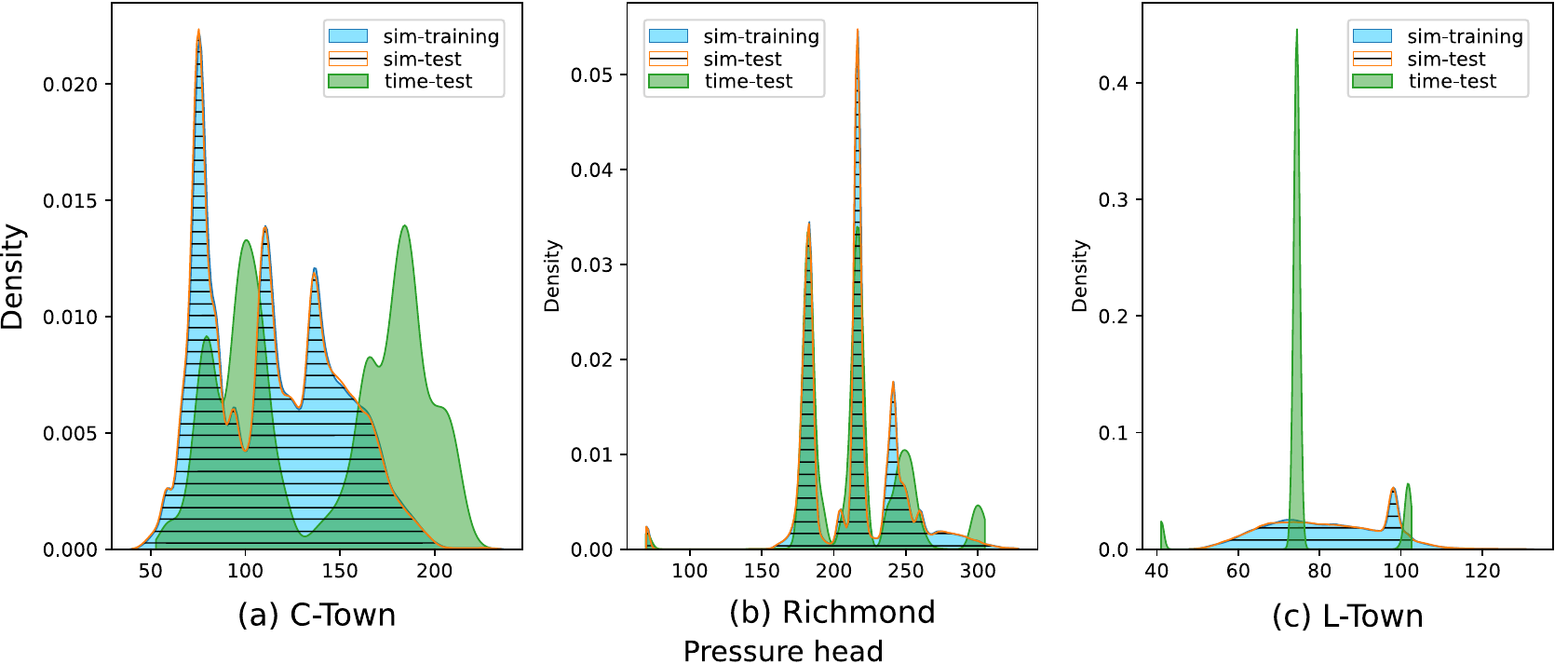}
		\caption{Comparison of density distributions of synthetic and time-based datasets in C-Town, Richmond and L-Town WDNs.}
		\label{fig:bsln_data_dist}
	\end{figure}

	The density distributions of training and test sets in C-Town, Richmond and L-Town WDNs are shown in Figure \ref{fig:bsln_data_dist}. It is clear that the distributions of the training and testing data created by the same algorithm (mathematical simulation) are identical, while the distribution of the test data with a demand pattern greatly differs from the one used during training. This shows the ability of \textit{GATRes} to adapt to the changes that occur in real life scenarios. It also shows that other models achieve better results only when evaluated on fidelity copies of the training data, caused by overfitting due to the large model complexity of the previous approaches. The density distribution plot in Figure \ref{fig:bsln_data_dist}(b) explains the good performance of mGCN on Richmond WDN in terms of MAPE and NSE (Table \ref{tab:bsln_comp_time_inv}), the time-based demand pattern test dataset has a similar distribution than the one used for training.  
	
\subsection{Ablation study}% Andres
\label{sec:ablation}

	The ablation study presented in this section evaluates the importance of the different components of \textit{GATRes} model architecture, and the effect of their removal or alteration on performance. In every run, a specific component is removed or altered, and \textit{GATRes} is restored to its original version before a new change is made. The different variants used in the ablation study are the following:
	
	\begin{itemize}
		\item[] \textbf{Without Residual Connections (woResCon)}. The residual connections used within each \textit{GATRes} Block are removed.
		\item[] \textbf{Without Mean Aggregation (woMeanAggr)}. The Mean Aggregation applied after the second convolution within each \textit{GATRes} Block is removed and the residual connection is added to the output of the second convolution.
		\item[] \textbf{Without Residual Connection and Without Mean Aggregation (woResCon-woAggr)}. Both, the residual connection and the mean aggregation are removed from the \textit{GATRes} Block.
		\item[] \textbf{Mean Aggregation Outside the Block (MeanAggrOut)}. Instead of applying a mean aggregation within each block, it is applied only once in the forward pass, after the output of the last \textit{GATRes} Block and before the final Linear layer.
		%\item[] \textbf{Without Gradient Clipping (woGradientClipping)}. The gradient clipping technique is not applied during the training process.
	\end{itemize}

	These experiments were performed by training \textit{GATRes-small} on the C-Town WDN. As can be seen in Table \ref{tab:ablation}, removing the Residual Connections produced the highest negative impact on model performance for all metrics. 
	
	\begin{table}[h]
		\caption{Ablation study of \textit{GATRes} evaluated on C-Town 24-hour time series data.}
		\centering
		\label{tab:ablation}
		\resizebox{0.75\columnwidth}{!}{%
			\begin{tabular}{l l l l}
				\toprule
				Variants & MAE ($\downarrow$) & MAPE ($\downarrow$) & NSE ($\uparrow$) \\
				\midrule
                    GATRes-small & \textbf{09.4860} \small{$\pm$0.1822}  & 0.0690 \small{$\pm$0.001}  & \textbf{0.8480} \small{$\pm$0.0075} \\
                    woMeanAggr & 09.7479 \small{$\pm$0.1444}  & \textbf{0.0686} \small{$\pm$0.0009}  & 0.8473 \small{$\pm$0.0046} \\
                    woResCon-woAggr & 10.0934 \small{$\pm$0.1644}  & 0.0735 \small{$\pm$0.0010}  & 0.8150 \small{$\pm$0.0062} \\
                    MeanAggrOut & 10.3735 \small{$\pm$0.1251}  & 0.0735 \small{$\pm$0.0006}  & 0.8333 \small{$\pm$0.0058} \\
                    woResCon & 11.5362 \small{$\pm$0.1697}  & 0.0815 \small{$\pm$0.0008}  & 0.7694 \small{$\pm$0.0089} \\
				\bottomrule
				%\multicolumn{2}{l}{$^{a}$Footnote text here.}
		\end{tabular}}
	\end{table}

\section{Discussion}
\label{sec:discussion}
% Impact of model depth
% Result- Benefit of synthesis data
% Shall we replace the mathematical simulation with deep learning models?

In this section, we generally discuss our findings and technical changes that affect our model in estimating pressures on Water Distribution Networks. We first review changes that made \textit{GATRes} versions outperform other baselines and their limitations. Then, we discuss the role of synthetic data and the relationship between hydraulic simulation and surrogate models. Finally, we address the question of generalizability in the context of our research.

\subsection{General findings and limitation}
\textit{GATRes} qualified all criteria for model assessment as in Section \ref{sec:crit_satis} and achieved pressure reconstruction with an average relative error of 7\% and an absolute error of 1.93 water column meters on a 95\% masking rate (see Table \ref{tab:oos_comparison}). We attribute its success primarily to the fundamental blocks and training strategy. These blocks update the weights of connections using nodal features and, therefore, relax the original topology in a given Water Distribution Network. This relaxation provides robustness and generalizability to \textit{GATRes} in uncertain conditions and across diverse network topologies, which may vary in size, headloss formula, and component configurations. Furthermore, \textit{GATRes} utilizes a random sensor replacement strategy, eliminating the need for time-consuming retraining when a new sensor is introduced in the future. For these reasons, both blocks within the architecture and training strategy sharpen a \textit{GATRes} as a highly reusable and sustainable solution for predicting pressures in numerous Water Distribution Networks.

However, it is essential to acknowledge the limitations when \textit{GATRes} comes to scale. The limit becomes apparent when comparing the larger and smaller versions of \textit{GATRes} in Tables \ref{tab:oos_comparison} and \ref{tab:oos_comparison_noisy}. The larger \textit{GATRes} eventually reaches a saturation point of performance and is surpassed by its smaller counterpart. The same finding is available in both GATConvWat variants. They likely originate from inherent issues in graph neural networks, such as over-smoothing and over-squashing, where nodes tend to propagate redundant information excessively \citep{di2023over}. While \textit{GATRes} employs residual connections that partially mitigate the over-smoothing and mainly contribute to the model performance, as shown in Section \ref{sec:ablation}, they are unable to eliminate this phenomenon completely \citep{kipf2017semisupervised}. To address these issues in the future, potential solutions may include exploring graph rewiring strategies and subgraph sampling techniques.

\subsection{Benefit of synthetic data}
Throughout our experiments, the integration of data generated by our innovative tool has proven to be a game-changer when it comes to training deep models. Those results not only achieve remarkable accuracy but also indicate the helpfulness of synthetic data, especially when sensor records or simulation parameters are restricted. Indeed, these common issues have been found in many public benchmarks, such as five reviewed water networks in Section \ref{sec:datasets}, due to the missing historical patterns and privacy issues. They have made reproducibility a persistent challenge in water network research.

As a solution, our data generation tool extends the limits in approaching these public networks without confidential matters. For practical purposes, the synthesized training set could involve as many cases as possible, reducing the risk of long-term incidents that may not have occurred in historical records. Thus, it boosts model robustness when dealing with unforeseen scenarios.

\subsection{Relationship between hydraulic simulations and \textit{GATRes}}
Yet, an intriguing question arises: Can we replace traditional mathematical simulations with surrogate models like \textit{GATRes}? Conventional simulation bridges the interaction between hydraulic experts and the Water Distribution Network in water management. Such an interaction should be preserved in the design or analysis phase. In the deployment, especially for Digital Twin or water systems on big data, pressure estimate models often deal with heavy computation and require a low response time \citep{pesantez2022digitaltwin}. In this case, \textit{GATRes} and GNN variants can be alternative approaches due to their competitive results and impressive throughput (see Table \ref{tab:throughput}). However, these deep models may involve the risk of over-relaxation of energy conservation laws and other constraints within the actual networks. The risk is often minimal in pure physics-based simulations.

Accordingly, these simulations still play a critical role in data synthesis as they define a valid boundary for newly created models thanks to their generated training samples and testing environments. When fast computation is required, \textit{GATRes} is a good alternative to estimate the pressure of a large WDN given unlimited sensor streams. In the future, it is potential to focus on physics-inspired models that can regularize \textit{GATRes} to preserve fundamental physical laws and yield more confident results.

\subsection{Generalization}
% generalization --> ANDRES
%Generalization in the context of our work is defined as the ability of the model to learn 

\textit{GATRes} is able to generalize to previously unseen WDNs by design given the ability of spatial methods (e.g. \textit{GAT}) to generalize across graphs \citep{bronstein2017geometric}. On the contrary, previous works that rely on spectral approaches suffer from the generalization problem because their convolutions (e.g. ChebNet) depend on the eigen-functions of the Laplacian matrix of a particular graph \citep{zhang2019graph}. 

\textit{GATRes} trained on multiple WDNs simultaneously produced a MAE of 3.06mH$_2$O and a MAPE of 9.98\%, on average, at zero-shot inference on 24-hour Oosterbeek WDN. These results are impressive given that pressure estimation was performed on a completely different, previously unseen, WDN. Moreover, the fine-tuned model, on the target dataset Oosterbeek, produced a reduction in MAE of 1.51\% with respect to the model trained directly and only on Oosterbeek. 

The results of our first attempts towards generalization (see Table \ref{tab:generalization}) show that our approach is worth further exploration. GNN models fail to generalize when the local structures of the graphs in the training data differ from the local structures in the test data \citep{yehudai2021local}. Then, a possible explanation of the generalization capabilities of \textit{GATRes} is the training on several WDNs simultaneously. Using graphs that differ in size and structure, for training, allows \textit{GATRes} to learn a richer set of local structures that may be present in the target WDN. Despite these promising results, several questions remain unanswered. For example, how to choose the right WDNs in order to enrich the training data in terms of local structures' diversity? How to  design a pre-trained task that can effectively capture the local-level patterns and extrapolate those to unseen larger graphs? How to train a GNN-based foundation model, in the Water Management Domain, that can be applied on different downstream tasks on any WDN topology? All these questions open paths for future research directions.

%\quickwordcount{discussion} 

% generalization --> ANDRES

\section{Conclusions}

    In this work we presented a hybrid, physics-based and data-driven, approach to address the problem of state estimation in WDNs. We leveraged mathematical simulation tools and GNNs to reconstruct the missing pressures at 95\% of the junctions in the network, from only 5\% of them seen during training. We also tested our approach on more extreme cases of sensor sparcity, reaching up to 99\% masking rate. Our work proposes a number of research contributions. First, a new training data generation process that does not consider time-dependent patterns and includes control parameters for the simulation that were fully overlooked in previous works. This results in a more diverse training dataset and avoids uncertainty propagation due to model simplification errors. In addition, our random masking strategy during training provides robustness against sensors' location changes due to new installations or maintenance. Moreover, the proposed evaluation method considers real time-dependent patterns and Gaussian noise injection, producing the out-of-distribution data intrinsic to real-world scenarios. Thus, enabling the resilience of the model to unexpected circumstances. Furthermore, a multi-graph pre-training strategy followed by fine-tuning allowed to improve the performance of the model with respect to the one trained and evaluated on a single topology. 
    
    % Our model was evaluated on a large-scale network in The Netherlands, Oosterbeek, as well as, on other WDNs benchmark datasets. The result on Oosterbeek, an average MAE of 1.94mH$_2$O, shows a reduction of 8.57\% with respect to other models. Similarly, the performance gain on other WDNs benchmarks goes from $\approx$5\% to $\approx$52\%, with respect to previous approaches. Most importantly, the evaluation of \textit{GATRes} under noisy conditions shows that our model is robust to uncertainties and unexpected changes in the system, which could not be observed on previous approaches. 

    Our model was evaluated on a large-scale network in The Netherlands, as well as on several WDNs benchmark datasets, showing a clear improvement over previous approaches. \textit{GATRes} obtained an average MAE of 1.94mH$_2$O, which represents an 8.57\% improvement with respect to other models. Similarly, it showed a reduction of MAE up to $\approx$52\% on other WDN benchmarks, in the best cases, with respect to previous approaches. We attribute the high performance of \textit{GATRes} to its building blocks and training strategy. These blocks relax the original topology leveraging nodal features to re-weight the connections by means of an attention mechanism. Despite its success, there are still some aspects that demand further exploration. On the one hand, while the residual connections mitigate the over-smoothing problem, inherent to GNNs, the phenomenon is not completely removed. Therefore, other techniques such as graph rewiring and subgraph sampling would be a fruitful area for further work. On the other hand, our multi-graph pre-training strategy is a promising direction towards model generalization and transferability in the WDNs domain. Nonetheless, further research needs to examine more closely the links between the topologies of the WDNs chosen for pre-training, the pre-training task, and their effect on the generalization capabilities of the model.

\label{sec:conclusion}

\section*{Data Availability Statement}
% This section MUST contain a statement that describes where the data supporting the conclusions can be obtained. Data cannot be listed as ''Available from authors'' or stored solely in supporting information. Citations to archived data should be included in your reference list. Wiley will publish it as a separate section on the paper’s page. Examples and complete information are here:
% https://www.agu.org/Publish with AGU/Publish/Author Resources/Data for Authors

In this section, we provide an overview of the publicly available benchmark water distribution networks and libraries that were employed in our study. Specifically, three networks, including Anytown \citep{walski1987battle}, C-Town \citep{ostfeld2012battle}, and Richmond \citep{van2001methodology}, have been collected on Github \citep{hajgato2021repository}. The L-town dataset is referenced in the paper by \citep{Vrachimis2022}, while the Ky13 benchmark \citep{hernadez2016water} can be readily obtained via a free download on \url{https://www.uky.edu/WDST/database.html}. The ``Large" network is referenced in the ``Availability of Data and Materials" section in \citep{sitzenfrei2023dual}. The Oosterbeek water network is not publicly available, as it is provided under confidentiality by the water provider Vitens.

In terms of libraries, we employed Matplotlib version 3.7.1 \citep{hunter2007matplotlib} (licensed under BSD) and Plotly version 5.15 \citep{plotly2015plotly}, licensed under MIT, to craft our visual figures. The data generation tool was constructed using the Epynet wrapper, available on \url{https://github.com/Vitens/epynet}, and is licensed under Apache-2.0. We also leveraged the WNTR library \citep{klise2018overview}, and Ray version 2.3.1 \citep{moritz2018ray} in our implementation. 

Our datasets are organized in the \textit{zarr} format, a file storage structure created using the Zarr-Python package version 2.14.2 \citep{miles2020zarr}, which is licensed under MIT. These datasets were employed in training both baseline models and \textit{GATRes} variants using PyTorch \citep{paszke2019pytorch} and PyTorch Geometric \citep{fey2019pyg}. We also tracked the training using tracking tools such as Wandb \citep{wandb2020wandb} and Aim \citep{Arakelyan2020Aim}. The dataset generation tool and \textit{GATRes} models are open-sourced at \url{https://github.com/DiTEC-project/gnn-pressure-estimation}.

\section*{Acknowledgments}
% Enter acknowledgments here. This section is to acknowledge funding, thank colleagues, enter any secondary affiliations, and so on.
This work is funded by the project DiTEC: Digital Twin for Evolutionary Changes in Water Networks (NWO 19454). We express our appreciation to Ton Blom and the Digital Twin group at Vitens, a Dutch drinking water company, for providing hydraulic knowledge and valuable data. Furthermore, we are grateful to Prof. A. Veldman for our insightful discussions. Also, we thank the Center for Information Technology of the University of Groningen for their support and for providing access to the Hábrók high performance computing cluster.  We also thank M. Hadadian, F. Blaauw and Researchable for discussions about the experiments platform. 

\bibliographystyle{apalike} 
\bibliography{references}

\end{document}